\documentclass[twoside,journal]{IEEEtran}

\usepackage{color}
\usepackage{cite}
\usepackage{amsmath}
\usepackage{amssymb}
\usepackage{pifont}
\newcommand{\cmark}{\ding{51}}%
\newcommand{\xmark}{\ding{55}}%
\usepackage{graphicx}
\usepackage{epstopdf}
\usepackage{tabularx}
\usepackage{booktabs} 
\usepackage[usenames,dvipsnames]{xcolor}
\usepackage{eso-pic}
\usepackage{textcomp}
\usepackage{pifont}
\usepackage{anyfontsize}
\usepackage{changepage}
\usepackage{setspace}
\usepackage{url}
\usepackage{multirow,bigdelim,makecell}
\usepackage{arydshln}
\usepackage{framed}

\usepackage{hyperref}

\ifCLASSINFOpdf
\else
\fi

\AddToShipoutPicture*{\small \raisebox{26.7cm}
{\hspace*{2.2cm}
\fboxrule=1.5pt
\fboxsep=0.0pt
\fbox{\colorbox{pink}
{
\begin{minipage}[c]{0.95\textwidth}
\footnotesize
{\setlength{\baselineskip}{0.5\baselineskip}
\vspace*{1pt}
This document is the paper as accepted for publication in \textit{Proceedings of the IEEE} (doi:\href{https://doi.org/10.1109/JPROC.2023.3273520}{10.1109/JPROC.2023.3273520}).\newline
\textcopyright 2023 IEEE.  Personal use of this material is permitted.  Permission from IEEE must be obtained for all other uses, in any current or future media, including reprinting/republishing this material for advertising or promotional purposes, creating new collective works, for resale or redistribution to servers or lists, or reuse of any copyrighted component of this work in other works.
\par}
\end{minipage}}}}}%

\begin{document}

\title{\vspace*{-3mm}Bottom-Up and Top-Down Approaches for the Design of Neuromorphic Processing Systems: Tradeoffs and Synergies Between\\ Natural and Artificial Intelligence}
\author{Charlotte~Frenkel,~\IEEEmembership{Member,~IEEE,}
        David~Bol,~\IEEEmembership{Senior Member,~IEEE,}
        and~Giacomo~Indiveri,~\IEEEmembership{Senior Member,~IEEE}\vspace*{-8mm}
\thanks{Manuscript received xxxxx, 2022; revised xxxxx, 2022. This work was supported in part by the CHIST-ERA grant CHIST-ERA-18-ACAI-004 (SNSF 20CH21186999 / 1), the European Research Council (ERC) under the European Union's Horizon 2020 research and innovation program grant agreement No 724295, the fonds europ\'een de d\'eveloppement r\'egional FEDER, the Wallonia within the ``Wallonie-2020.EU'' program, the Plan Marshall, and the National Foundation for ~\mbox{Scientific Research (F.R.S.-FNRS) of Belgium.}}
\thanks{\textbf{C. Frenkel} was with the Institute of Neuroinformatics, University of Zurich and ETH Zurich, Zurich CH-8057, Switzerland. She is now with the Faculty of Electrical Engineering, Mathematics and Computer Science (EEMCS), Department of Microelectronics, Delft University of Technology, 2628 CD Delft, The Netherlands (e-mail: c.frenkel@tudelft.nl).}
\thanks{\textbf{G. Indiveri} is with the Institute of Neuroinformatics, University of Zurich and ETH Zurich, Zurich CH-8057, Switzerland (e-mail: giacomo@ini.uzh.ch).}
\thanks{\textbf{D. Bol} is with the ICTEAM Institute, Universit\'e catholique de Louvain, Louvain-la-Neuve BE-1348, Belgium (e-mail: david.bol@uclouvain.be).}%
\thanks{Digital Object Identifier 10.1109/JPROC.2023.3273520}}%

\markboth{Frenkel \MakeLowercase{\textit{et al.}}: Bottom-Up and Top-Down Neuromorphic Systems}%
{Frenkel \MakeLowercase{\textit{et al.}}: Bottom-Up and Top-Down Neuromorphic Systems}%

\maketitle
\thispagestyle{empty}

\begin{abstract} 

While Moore's law has driven exponential computing power expectations, its nearing end calls for new avenues for improving the overall system performance. One of these avenues is the exploration of alternative brain-inspired computing architectures that aim at achieving the flexibility and computational efficiency of biological neural processing systems. Within this context, neuromorphic engineering represents a paradigm shift in computing based on the implementation of spiking neural network architectures in which processing and memory are tightly co-located.
  In this paper, we provide a comprehensive overview of the field, highlighting the different levels of granularity at which this paradigm shift is realized and comparing design approaches that focus on replicating natural intelligence (bottom-up) versus those that aim at solving practical artificial intelligence applications (top-down). First, we present the analog, mixed-signal and digital circuit design styles, identifying the boundary between processing and memory through time multiplexing, in-memory computation, and novel devices. Then, we highlight the key tradeoffs for each of the bottom-up and top-down design approaches, survey their silicon implementations, and carry out detailed comparative analyses to extract design guidelines. Finally, we identify necessary synergies and missing elements required to achieve a competitive advantage for neuromorphic systems over conventional machine-learning accelerators in edge computing applications, and outline the key ingredients~\mbox{for a framework toward neuromorphic intelligence.}

\end{abstract} 


\IEEEpeerreviewmaketitle

\vspace*{-3mm}
\section{Introduction} \label{sec_intro}

\IEEEPARstart{T}OGETHER with the development of the first mechanical computers came the ambition to design machines that can think, with first essays dating back to 1949~\cite{Berkeley49,Turing50}. The advent of the first silicon computers in the 1960s, together with the promise for exponential transistor integration (i.e.~\textit{Moore's law}, as first coined by Carver Mead~\cite{Gelsinger06}), further fuelled that ambition toward the development of embedded artificial intelligence. As a key step toward brain-inspired computation, artificial neural networks (ANNs) were introduced based on the observation that the brain processes information with densely-interconnected and distributed computational elements: the neurons. The successful deployment of the backpropagation of error (BP) learning algorithm, backed by steep progress in CPU and GPU computing resources, recently enabled a massive scaling of ANNs, allowing them to outperform many classical optimization and pattern recognition algorithms~\cite{Rumelhart86,Schmidhuber15}.
Today, deep neural networks form a significant part of \textit{artificial intelligence}~(AI) research~\cite{LeCun15}, with applications ranging from machine vision (e.g.,~\cite{Krizhevsky12,LeCun15,He16}) to natural language processing~(e.g.,~\cite{Hinton12,Amodei16,Brown20}), often nearing or outperforming humans in complex benchmarking datasets, games of chance and even medical diagnosis~\cite{He15,Moravcik17,Olczak17}. Yet, most of these AI successes focus on specialized problem areas and tasks, which can be referred to as \textit{narrow AI}~\cite{Goertzel07}. Although recent efforts aim at the development of an artificial intelligence~that is both more general and multi-modal~\cite{Goertzel07,Chollet19,LeCun22,Zador22,GPT4}, current application-specific AI solutions deployed on centralized computing backends show a lack of both \textit{versatility} and \textit{efficiency} when compared to biological brains.

\paragraph*{The versatility gap}~Despite the wide diversity of the above-mentioned applications, task versatility is limited as each use case requires a dedicated and optimized network. Porting such networks to new tasks would at best require retraining with new data, and at worst imply a complete redesign of the neural network architecture, besides retraining. The need to tailor and retrain networks for each use case is problematic as the amount of both data and computation needed to tackle state-of-the-art complex tasks has been growing by an order of magnitude approximately every year in the last decade. This growth rate is much faster than that of technology scaling, and outweighs the efforts to reduce the network computational footprint~\cite{Thompson20}. To improve the ability of ANN-based AI to scale, diversify, and generalize from limited data while avoiding catastrophic forgetting, few-shot learning approaches based on meta-learning techniques are being investigated~\cite{Schmidhuber87,Thrun98,Riemer19,Hospedales20,Henning21,Zucchet21}.  These approaches aim at building systems that are tailored to their environment and can quickly adapt once deployed, just as evolution shapes the degrees of versatility and online adaptation of biological brains~\cite{Wang21a}. These are key aspects of the human brain, which excels at learning a model of the world from few examples~\cite{Hawkins19}.

\paragraph*{The efficiency gap}~For tasks that animals need to solve, such as sensory processing, classification, or pattern recognition, the power and area efficiencies of current AI systems lag behind biological ones at all levels of complexity. Taking the game of Go as a well-known proxy for complex applications, both task performance and efficiency ramped up quickly. From AlphaGo Fan~\cite{Silver16}, the first computer to defeat a professional player, to AlphaGo Zero~\cite{Silver17a}, the one now out of reach from any human player, power consumption went from 40\,kW to only about 1\,kW~\cite{Silver17b}.
However, even in its most efficient version, AlphaGo still lags two orders of magnitude behind the \mbox{20-W} power budget of the human brain. While most of this gap could potentially be recovered with a dedicated hardware implementation, AlphaGo would still be limited to a single task. On the other end of the spectrum, for low-complexity tasks, a centralized cloud-based AI approach is not suitable to endow resource-constrained distributed wireless sensor nodes with intelligence, as data communication would dominate the power budget~\cite{Bol15}.
The trend is thus shifting toward decentralized near-sensor data processing, i.e.~\textit{edge computing}~\cite{Shi16}. Shifting processing to the edge requires the development of dedicated hardware accelerators tailored to low-footprint ANN architectures, recently denoted as tinyML~\cite{LeCun19,Sylvester20,Banbury20}.
However, state-of-the-art ANN accelerators currently burn on the order of milliWatts for basic image classification on small pixel patches\footnote{As for the CIFAR-10 dataset~\cite{Krizhevsky09}, comprising 10 classes of animal and vehicle images in a format of 32 by 32 pixels. Hardware accelerator from~\cite{Bankman19} taken as a reference.}, thereby still lagging orders of magnitude behind biological efficiency. As a point of comparison, the honey bee brain has about one million of neurons for a power budget of 10\,$\mu$W only, yet it is able to perform tasks ranging from real-time navigation to complex pattern recognition, while constantly adapting to its environment~\cite{Sandin14}.
In order to minimize the energy footprint of edge computing devices, state-of-the-art techniques include minimizing accesses to centralized memories~\cite{Sze17} and in-memory computing~\cite{Verma19}, advanced always-on wake-up controllers~\cite{Giraldo20,An21}, as well as weight and activation quantization~\cite{Courbariaux16,Hubara18}. The field is thus naturally trending toward key properties of biological neural processing systems: processing and memory co-location, event-driven processing, and low-precision computation with a binary spike encoding, respectively.

\begin{figure}[!t]
\centering
\noindent\includegraphics[width=1.00\columnwidth]{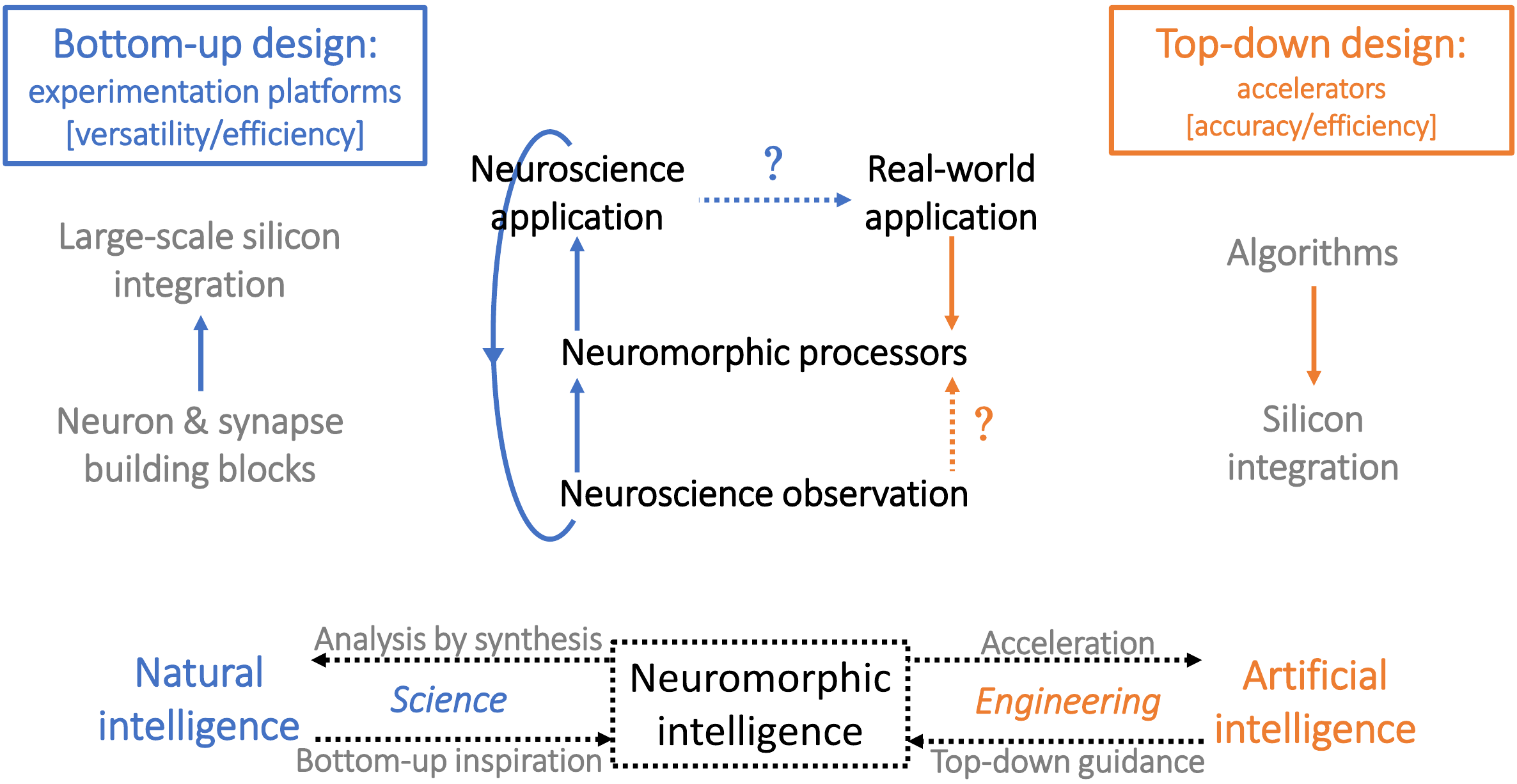}
\caption{Summary of the bottom-up and top-down design approaches toward neuromorphic intelligence. Bottom-up approaches optimize a tradeoff between versatility and efficiency; their key challenge lies in stepping out from analysis by synthesis and neuroscience-oriented applications toward demonstrating a competitive advantage on real-world tasks. Top-down approaches optimize a tradeoff between task accuracy and efficiency; their key challenge lies in optimizing the selection of bio-inspired elements and their abstraction level. Each approach can act as a guide to address the shortcomings of the other.}%
\vspace*{-1mm}
\label{fig_strategy}
\end{figure}

\begin{figure*}[!t]
\centering
\noindent\includegraphics[width=0.963\textwidth]{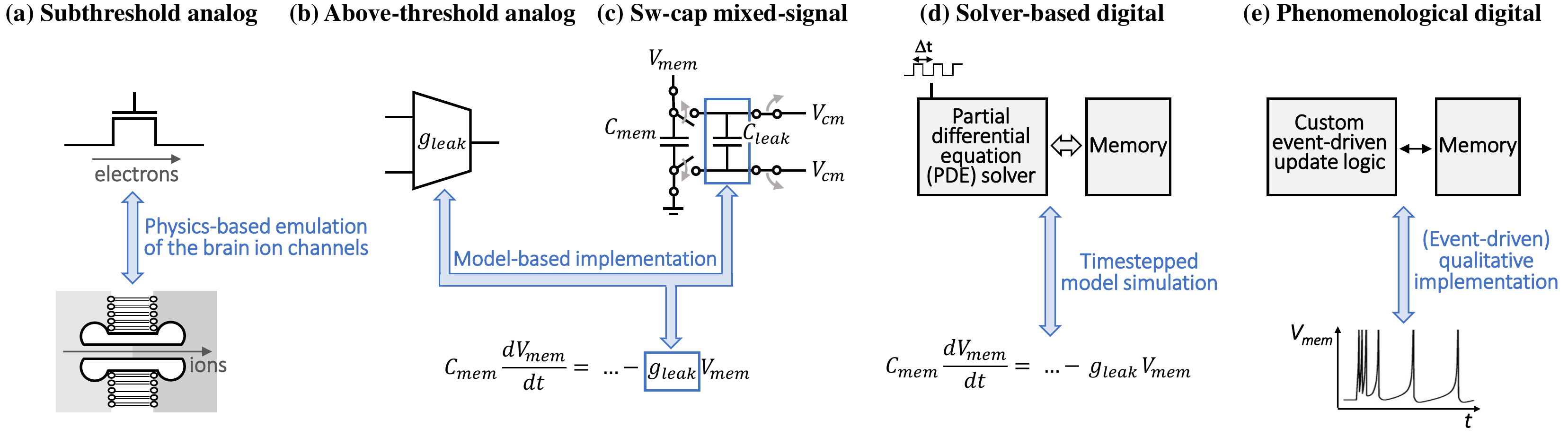}
\caption{Overview of the different neuromorphic circuit design styles, together with their key implementation strategies. \textbf{(a)}~Subthreshold analog design offers a direct emulation of the brain ion channel dynamics directly grounded on the device physics of the MOS transistor, as the ion (resp. electron) flow in the brain ion channels (resp.~the channels of MOS transistors in subthreshold regime) is governed by a \textit{diffusion} mechanism. \textbf{(b-c)}~Above-threshold analog design and switched-capacitor mixed-signal design rely on a circuit implementation that corresponds one-to-one with the selected neuron mathematical model. \textbf{(d)}~Solver-based digital design offers a straightforward approach based on partial differential equation (PDE) solvers, which can solve the chosen neuron mathematical model. This comes at the expense of time-discretization and large data movement, given that the PDE solver has to fetch/update from/to memory the model state at discrete mathematical integration timesteps $\Delta$t. \textbf{(e)}~These penalties can be reduced by following a phenomenological digital design approach, which implements specific neuron behaviors qualitatively using custom update logic, which can accommodate for sparse event-driven updates instead. Note that figures (b-e) focus on neuron models, but all design strategies~\mbox{can be applied to synapse or other biological computational primitives without loss of generality.}}%
\label{fig_designStyles}
\end{figure*}

\begin{table*}
\vspace*{-1mm}
\caption{Properties and tradeoffs of the different neuromorphic circuit design styles.\\%
Elements usually representing key design drivers are highlighted in bold.}
\vspace*{-1mm}
\label{tab_design}
\centering
\resizebox{0.90\textwidth}{!}{
\begin{tabular}{lccccc}
\toprule%
Implementation & \multicolumn{2}{c}{Analog} & Mixed-signal & \multicolumn{2}{c}{Digital} \\
 & Subthreshold & Above-threshold & Switched-capacitor & Solver-based & Phenomenological \\
\midrule%
Dynamics      & \textbf{Physics-based} & Model-based & Model-based & Timestepped & Event-driven$^*$ \\
Versatility/efficiency tradeoff & \textbf{(Excellent)$^\ddag$} & Medium & \textbf{Good} & Bad & \textbf{Good} \\
Time constant & \textbf{Biological} & \textbf{Accelerated} & Biological to accelerated & \multicolumn{2}{c}{Biological to accelerated} \\
Noise, mismatch, PVT sensitivity & High & Medium & \textbf{Medium to low} & \multicolumn{2}{c}{--} \\
\multirow{2}{*}{Indirect overhead} & \multicolumn{2}{c}{\multirow{2}{*}{Bias generation}} & \multirow{2}{*}{Clocked digital control} & \multicolumn{2}{c}{\multirow{2}{*}[-1pt]{\shortstack{Clock tree (sync)\\ Low tool support (async)}}} \\
&&&&\\
\multirow{2}{*}{Design time} & \multicolumn{2}{c}{\multirow{2}{*}{High}} & \multirow{2}{*}{High} & \multicolumn{2}{c}{\multirow{2}{*}[-1pt]{\shortstack{\textbf{Low (sync)}\\\textbf{Medium (async)}}}} \\
&&&&\\
Technology scaling potential & \multicolumn{2}{c}{Low} & \textbf{Medium} & \multicolumn{2}{c}{\textbf{High}} \\
&&&&\\
Programmability & \multicolumn{2}{c}{Low} & Low & \multicolumn{2}{c}{\textbf{High}} \\
\bottomrule%
\end{tabular}}%
\flushleft
\scriptsize

\hspace*{1.1cm}$^*$ Although phenomenological digital designs can also implement timestepped updates, event-driven updates are the preferred choice to reduce data movement.

\hspace*{1.1cm}$^\ddag$ Degrades at the system level if variability is not exploited and requires compensation.
\end{table*}

Therefore, to reach the goal of versatile and efficient computing electronic technologies, taking biological brains as a guide appears as a natural research direction.
This strategy started in the late 1980s with neuromorphic engineering. The term ``neuromorphic'' was coined by Carver Mead with the observation that direct emulation of the brain ion channels dynamics could be performed by the MOS transistor operated in the subthreshold regime~\cite{Mead89}. The field of neuromorphic engineering lies at the crossroads of neuroscience, computer science, and electrical engineering. It encompasses the study and design of bio-inspired systems following the biological \textit{organizing principles} and \textit{information representations}. Therefore, at least in principle, the field of neuromorphic engineering aims at a two-fold paradigm shift. Firstly, while conventional von Neumann processor architectures rely on separated processing and memory, the brain organizing principles rely on distributed computation that co-locates processing and memory with neuron and synapse elements, respectively~\cite{Indiveri15a}. This first paradigm shift therefore aims at releasing the von Neumann bottleneck in data communication between processing and memory, a point whose criticality is further emphasized by the recent slow down in the pace of Moore's law, especially for off-chip DRAM memory~\cite{Horowitz14}. Secondly, conventional von Neumann processor architectures encode data as multi-bit words that are processed sequentially by instructions, orchestrated by a global clock. Time is thus a by-product of computation and the resolution is determined by the number of bits used for encoding. On the contrary, the brain processes information by encoding data both in space and time with all-or-none binary spike events, each single axon potentially encoding arbitrary precision in the inter-spike time interval~\cite{Rieke96,Thorpe01}, where \emph{time represents itself}. This second paradigm shift aims at sparse event-driven processing toward a reduced power consumption, especially if spikes are used all the way from sensing to computation. However, these paradigm shifts are often not fully attained in actual neuromorphic hardware: the granularity at which they are realized depends on the implementation choices and the design strategy that is followed, the latter being of two types: either \textit{bottom-up} or \textit{top-down} (see Fig.~\ref{fig_strategy}).%

The former design strategy takes neuroscience as the starting point: it is a \textit{basic research} approach toward understanding \textit{natural intelligence}, backed by the design of experimentation platforms optimizing a tradeoff between the \textit{versatility} of the biophysical behaviors that can be reproduced and the system-level \textit{efficiency} (i.e.~versatility/efficiency tradeoff). The latter one departs from the selected use case: it is an \textit{applied research} approach grounded on today's ANN successes toward solving \textit{artificial intelligence} applications, backed by the design of dedicated hardware accelerators optimizing a tradeoff between the task-level \textit{accuracy} and the system-level \textit{efficiency} (i.e.~accuracy/efficiency tradeoff). At the crossroads of both approaches, we argue that \emph{neuromorphic intelligence} can form a unifying substrate toward the design of low-power bio-inspired neural processing systems.
Extending from~\cite{FrenkelThesis20}, this paper surveys key design choices and implementation strategies, thereby complementing previous circuit-, algorithm- or system-level reviews~\cite{Schuman17,Thakur18,Bouvier19,Roy19,Basu22}. We will first cover the different styles of analog and digital design, together with tradeoffs brought by time multiplexing and novel devices (Section~\ref{sec_design}). Next, we will survey bottom-up design approaches in Section~\ref{sec_bottomup}, from the building blocks to their silicon implementations. We will then survey top-down design approaches in Section~\ref{sec_topdown}, from the algorithms to their silicon implementations. For both bottom-up and top-down implementations, detailed comparative analyses will be carried out so as to extract key insights and design guidelines. Finally, in Section~\ref{sec_discussion}, we will outline the key synergies between both approaches, the open challenges and the perspectives toward on-chip neuromorphic intelligence for autonomous agents that efficiently and continuously adapt to their environment.

\section{Neuromorphic circuit design styles} \label{sec_design} \vspace*{3mm}

Regardless of the chosen bottom-up or top-down approach to the design of neuromorphic systems, different circuit design styles can be adopted, as shown in Fig.~\ref{fig_designStyles}. Usually, a key question consists in choosing whether an analog or a digital circuit design style should be selected. In this section, we provide a principled analysis for choosing the circuit design style that is appropriate for a given use case.

Analog and digital neuromorphic circuit design each come in different flavors with specific tradeoffs. A qualitative overview is provided in Table~\ref{tab_design}. The tradeoffs related to analog and mixed-signal design are analyzed in Section~\ref{ssec_design_analog}, and those of digital design in Section~\ref{ssec_design_digital}. Important aspects related to memory and computing co-location, such as time multiplexing and in-memory computation, are discussed in Section~\ref{ssec_design_tmux}. This highlight of the key drivers behind each circuit design style is then illustrated in Sections~\ref{sec_bottomup} and~\ref{sec_topdown}, where actual neuromorphic circuit implementations are presented and compared.
\vfill

\subsection{Analog and mixed-signal design}\label{ssec_design_analog}

\textit{Subthreshold}, or \emph{weak-inversion} analog circuit design (Fig.~\ref{fig_designStyles}(a)) allows leveraging an \textit{emulation} approach directly grounded on the physics of the silicon substrate. Indeed, in the subthreshold regime, the current flow in the MOS transistor channel is governed by a diffusion mechanism, which is the same mechanism as for the ion flow in the brain ion channels~\cite{Mead89}. This emulation approach allows for the design of compact and low-power neuromorphic circuits that lie close to the brain biophysics. Considering voltage swings of 1\,V for capacitors and currents on the order of 1pF and 1nA, respectively, the resulting time constants are on the order of milliseconds~\cite{Indiveri11}, close to those observed in biology. Subthreshold analog designs are thus inherently adapted for real-time and closed-loop processing of natural signals, using time constants that are well-matched to those of environmental and biological \mbox{stimuli}.
Therefore, device-level biophysical modeling makes subthreshold analog designs suited for efficient brain emulation and basic research through analysis by synthesis. Subthreshold analog design allows for the emulation of a large range of neuronal behaviors and synaptic dynamics with few transistors, which we denote as an excellent \textit{versatility/efficiency} tradeoff at the building-block level, i.e.~individual neurons and synapses. However, these circuits are characterized by high sensitivity to noise, mismatch, and power, voltage and temperature (PVT) variations. Ensuring reliable computation at the system level thus requires applying circuit calibration procedures~\cite{LeneroBardallo08,Neftci10,KA17} or increasing redundancy in neuronal resources so as to combine robust computational primitives~\cite{Neckar19,Liang19,Zendrikov22}. Although these \textit{compensation} techniques currently appear to degrade the versatility/efficiency tradeoff at the system level, they might provide additional benefits if variability can be \textit{exploited} for computation and learning. Recent trends include variability-aware training~(see Section~\ref{ssec_algos}) and exploiting neural parameter variability to support efficient and robust learning with temporal data~\cite{Lengler_etal13, Mahvash13, Balasubramanian15, PerezNieves21, Zeldenrust21, Whittington_etal22}.%

\textit{Above-threshold} analog design (Fig.~\ref{fig_designStyles}(b)) is suited for accelerated-time modeling of biological neural networks. Indeed, compared to subthreshold analog designs, even when the capacitor size is of the same order (e.g., of 1\,pF), higher currents and reduced voltage swings produce  acceleration factors ranging from $10^3$ to $10^5$ compared to biological time, thus mapping year-long biological developmental timescales to day-long runtimes~\cite{Schemmel07,Schemmel10,Schemmel20}. However, as the current flow in the channel of the MOS transistor operated in the above-threshold regime is governed by a drift mechanism instead of diffusion, emulation of neural processes cannot take place anymore at the level of the device physics. Instead, the implementation of neural processes is done at a higher level by following the selected neuron/synapse mathematical model: following a structured analog design approach, appropriate analog circuits with tunable parameters are designed for each term of the equations in the chosen models~\cite{Aamir18a}. Although transistors operated in the above-threshold regime have an improved robustness to noise, mismatch and PVT variations compared to the ones operated in subthreshold, device mismatch is still a critical problem that requires mitigation at the circuit and system levels. Therefore, calibration procedures are also common, and sometimes directly integrated in the hardware~\cite{Aamir18b}. 

Designs based on \textit{switched-capacitor} (SC) circuits (Fig.~\ref{fig_designStyles}(c)) exhibit an interesting blend between specific properties of sub- and above-threshold analog designs. Similarly to above-threshold designs, they follow a higher-level implementation, however computation is carried out in the charge domain instead of the current domain. SC neuromorphic designs are thus able to achieve not only accelerated time constants, but also biologically-realistic ones. Furthermore, replacing nanoampere-scale currents by the equivalent accumulated charge has the advantage of reducing the sensitivity to noise, mismatch and PVT variations~\cite{Folowosele11,Mayr16}. The price to pay, however, is the overhead added by the clocked digital control of SC circuits, which can take up a significant portion of the system power consumption. As the digital part of this overhead can benefit from technology scaling, an overall good versatility/efficiency tradeoff for SC circuits in advanced technology nodes is possible~\cite{Mayr16}. Switched capacitors can also be used to implement time multiplexing (see Section~\ref{ssec_design_tmux}).

\subsection{Digital design}\label{ssec_design_digital}

As opposed to their analog counterparts, digital designs forgo the emulation approach. Instead, they \textit{simulate} neural processes, thereby relying on circuit implementations that lie far from the biophysics, which does not allow exploiting the dynamics of the silicon substrate. More circuit resources are thus needed to reproduce a large repertoire of neural behaviors and synaptic dynamics, thereby degrading the versatility/efficiency tradeoff. In exchange, digital designs are robust to noise, mismatch and PVT variations, can leverage technology scaling, and can offer high programmability with the support of different models and functions. The former ensures a predictable behavior and possibly a one-to-one correspondence with the simulation software, while the latter ensures competitive power and area efficiencies with deep sub-micron technologies.

The most straightforward starting point for digital neuromorphic design is to implement \textit{solvers} for the partial differential equations (PDEs) modeling the biophysical behavior of neurons and synapses, which requires retrieving and updating all model states at every integration timestep~\cite{Cassidy13b,Luo15,Levi18,Yang18} (Fig.~\ref{fig_designStyles}(d)). This implies an extensive and continuous amount of data movement and computation, including when no relevant activity is taking place in the network. Therefore, these approaches have poor power and area efficiencies, especially at accelerated time constants. Piecewise linear approximations of neuron models have been proposed to reduce the complexity and resource usage~\cite{Soleimani12,Yang15}, however they still require an update of all model states after each discrete mathematical integration timestep of the PDEs. In order to minimize updates, some studies analyzed the maximum integration timestep values for a given neuron model~\cite{Gunasekaran19}. In any case, the extensive data movement implied by solver-based digital implementations makes them difficult to match with a low-power event-driven neuromorphic approach.

\textit{Phenomenological} digital design (Fig.~\ref{fig_designStyles}(e)) aims at reducing the timestepped data movement overhead of its solver-based counterpart by carrying out updates when and where relevant in the neural network. To do so, two strategies can be followed: either the detail level of biophysical modeling can be reduced and the model simplified, or key behaviors of complex models can be qualitatively implemented using custom update logic, thereby forgoing the underlying mathematical model and the exact dynamics. While referring to Section~\ref{sssec_neurons} for the neuron models mentioned below, key examples on each side can be seen in:
\begin{itemize}
\item for the former, the popular leaky integrate-and-fire (LIF) neuron model, which eliminates all biophysical details of ion channels and only keeps the leaky integration property of the neuron membrane,
\item for the latter, the design of~\cite{Frenkel17a} that sidesteps the Izhikevich neuron model equations and instead aims at a low-cost reproduction of its firing behaviors.
\end{itemize}
In both examples given above, the model requirements are sufficiently relaxed so as to allow for event-driven state updates, thus strongly reducing data movement and the associated overhead. The strategy to be pursued and the approximations that can be made depend on the chosen application, therefore phenomenological digital design is a \textit{co-design} approach trading off model complexity, biophysical accuracy and implementation efficiency.

Finally, for both the solver-based and phenomenological approaches, a significant source of overhead is the clock tree, which for modern synchronous digital designs represents 20--45\% of the total power consumption~\cite{Sitik16}. Although clock gating techniques can help, this leads to a tradeoff between power and complexity that is a severe issue for neuromorphic circuits, whose activity should be event-driven. \textit{Asynchronous} digital circuits avoid this clock tree overhead and ideally support the event-driven nature of spike-based processing. This is the reason why asynchronous logic is a widespread choice for the on- and off-chip spike communication infrastructures of neuromorphic systems, both analog and digital. However, asynchronous circuit design currently suffers from a lack of native support in standard industrial computer-aided design (CAD) tools. Indeed, all neuromorphic systems embedding asynchronous logic rely on a custom tool flow (e.g.,~see~\cite{Painkras13,Benjamin14,Akopyan15,Davies18,Moradi18,Neckar19}), which increases the design time and requires support from a team experienced in asynchronous logic design. The custom flows employed in these designs all derive from the asynchronous digital design tools initially developed at Caltech in the 1990s~\cite{Manohar97}, which are now mainly maintained at Yale University and have recently been made open-source~\cite{Ataei21}. Another emerging solution consists in applying specific constraints to standard industrial digital CAD tools so as to automatically optimize the timing closure of asynchronous bundled-data circuits~\cite{Gibiluka15,Miorandi17,Gimenez18}. This idea was recently applied in the context of network-on-chips (NoCs), where Bertozzi~\textit{et~al.} demonstrate significant power-performance-area improvements for asynchronous NoCs compared to synchronous ones, while maintaining an automated flow based on standard CAD tools~\cite{Bertozzi21}. Leveraging the efficiency of asynchronous circuits with a standard digital tool flow may soon become a key element to support large-scale integration of neuromorphic systems.

\begin{figure}[!t]
\centering
\vspace*{1mm}
\noindent\includegraphics[width=0.75\columnwidth]{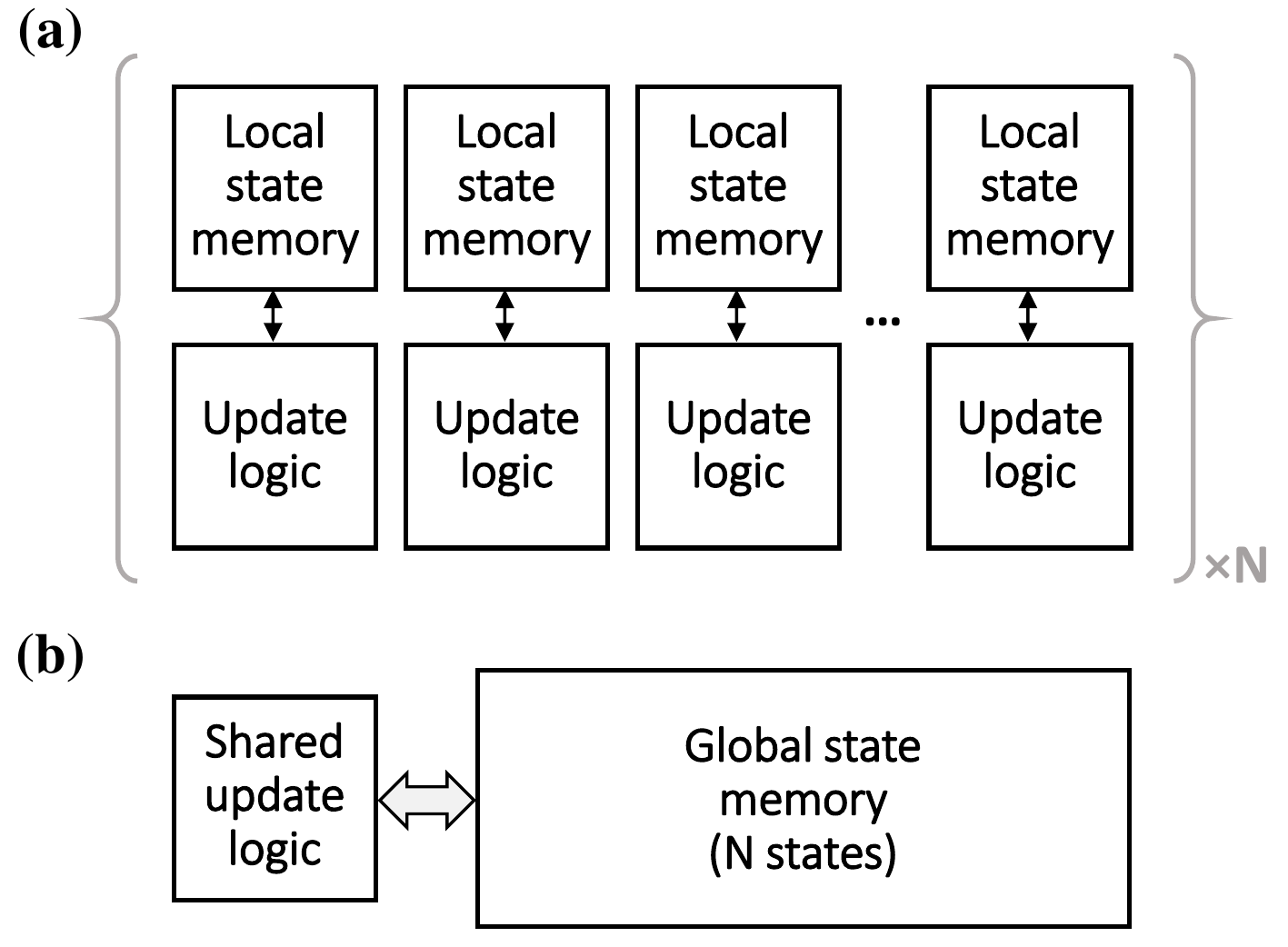}
\caption{Qualitative illustration of \textbf{(a)}~the fully-parallel and \textbf{(b)}~the time-multiplexed architectures for $N$ elements (e.g., neurons or synapses), and of their memory access bottleneck.}%
\vspace*{3mm}
\label{fig_tmux}
\end{figure}

\begin{table}
\caption{Properties and tradeoffs of fully-parallel and time-multiplexed designs. Elements usually representing key design drivers are highlighted in bold.}
\vspace*{-1mm}
\label{tab_tmux}
\centering
\resizebox{0.98\columnwidth}{!}{
\begin{tabular}{lcc}
\toprule%
Implementation & Fully-parallel & Time-multiplexed \\
\midrule%
\multirow{2}{*}{Time} & \multirow{2}{*}[-1pt]{\shortstack[l]{Analog: represents itself\\Digital: simulated}} & \multirow{2}{*}{Simulated} \\
&&\\
\multirow{2}{*}{Continuous dynamics} & \multirow{2}{*}{\textbf{Intrinsic} \cmark} & \multirow{2}{*}[-1pt]{\shortstack[l]{Timestepped updates: \cmark~(power $\uparrow$)\\Event-driven updates: \xmark}} \\
&&\\
\multirow{2}{*}{Mem/proc co-location} & \multirow{2}{*}{\textbf{Highest granularity}} & \multicolumn{1}{l}{\multirow{2}{*}[-1pt]{\shortstack[l]{SRAM: Cache-level granularity\\Off-chip DRAM: \xmark}}} \\
&&\\
Maximum throughput & High & Low \\
Power penalty & \textbf{Static} & \textbf{Dynamic} \\
Area footprint & High & \textbf{Low} \\
\bottomrule%
\end{tabular}}%
\end{table}

\vspace*{1mm}\subsection{Defining the boundary between memory and processing -- Time-multiplexing, in-memory computation and novel devices}\label{ssec_design_tmux}\vspace*{2mm}

Neuromorphic engineering aims at a paradigm shift from von-Neumann-based architectures to distributed and co-integrated memory and processing elements. However, the granularity at which this paradigm shift is achieved in practice strongly depends on the selected memory storage and on the level of resource sharing. Indeed, a key design choice for neuromorphic architectures consists in selecting between a fully-parallel resource instantiation and the use of a time multiplexing scheme (i.e.~shared update logic and centralized state storage), as shown in Figs.~\ref{fig_tmux}(a) and~\ref{fig_tmux}(b), respectively. A summary of the tradeoffs between both approaches is provided in Table~\ref{tab_tmux}. An important benefit of time multiplexing is the substantial reduction of the area footprint, usually by one to three orders of magnitude, at the expense of a reduction in the maximum throughput. This throughput reduction is usually not problematic, unless when targeting acceleration factors higher than one order of magnitude compared to biological time. Importantly, regarding the power consumption, the penalty for fully-parallel implementations is in static power (through the duplication of circuit resources with leakage power), while the penalty for time-multiplexed designs is in dynamic power (through an increase in memory accesses to a more centralized state storage). Therefore, minimizing leakage is necessary for fully-parallel designs, while state updates should be minimized for time-multiplexed ones, thereby highlighting the energy efficiency penalty of time-multiplexed PDE solvers carrying out updates at every integration timestep.

While time multiplexing based on on-chip SRAM memory is applied to nearly all digital designs due to its ease of implementation for a minimized area footprint, this technique is not applied to analog designs if a fully-parallel emulation of the network dynamics is to be maintained. Otherwise, time multiplexing can be applied to analog designs as well, as shown in~\cite{Schemmel10,Moradi13,Park14,Mayr16}. It can be either SRAM-based or capacitor-based, the former is a mixed-signal approach that minimizes the storage area for large arrays but requires digital-to-analog (DAC) converters, while the latter avoids DACs at the expense of a higher-footprint storage. In both cases, the addition of digital control logic is required. Furthermore, time multiplexing can also be applied selectively to different building blocks. As synapses are usually the limiting factor (Section~\ref{sssec_synapses}), a good example consists of time-multiplexed synapses and fully-parallel neurons, as in~\cite{Moradi13}, which represents an interesting tradeoff to minimize the synaptic footprint while keeping continuous parallel dynamics at the neuron level.

Finally, an important aspect of fully-parallel implementations is to enable synergies with \textit{in-memory computation}, where computation takes place in the memory itself, a trend that is popular not only in neuromorphic engineering~\cite{Christensen21}, but also in conventional machine-learning accelerators based on SRAM~\cite{Verma19}, DRAM~\cite{Seshadri17} and novel devices~\cite{Sebastian20}. A recent comparative analysis by Peng~\textit{et~al.} shows that, at normalized resolution and compared to six different memristor technologies, SRAM still offers the highest accuracy, throughput, density and power efficiency for deeply-scaled processes~\cite{Peng20}. However, while SRAM-based in-memory computation allows for efficient matrix-vector product acceleration, it is not typically encountered in spiking neural network (SNN) accelerators due to a lack of proper sparsity support, as opposed to fully-parallel memristor arrays.

Instead, fully-parallel \textit{memristor crossbar arrays} are a promising avenue for in-memory computation in neuromorphic systems~\cite{Payvand19,Mehonic20,Chicca20}. Beyond the usual prospects for improvement in density and power efficiency linked with in-memory computation, memristors offer specific synergies for neuromorphic engineering with characteristics similar to those of biological synapses~\cite{Indiveri13}, e.g.~learning dynamics, stochastic readout, few-bit device resolution, and dense nanoscale integration. Furthermore, a neuromorphic approach exploiting non-idealities instead of mitigating them could be particularly appropriate to alleviate the high levels of noise and mismatch encountered in these devices~\cite{Payvand19}, or to take advantage of parasitic effects such as the conductance drift~\cite{Demirag21}. However, high-yield large-scale co-integration with CMOS is still at an early stage~\cite{Lin14,Rofeh15}.

\vspace*{1mm}\section{Bottom-up design approach -- Trading off biophysical versatility and efficiency} \label{sec_bottomup}\vspace*{2mm}

The vast majority of neuromorphic designs follow a \textit{bottom-up} strategy, which is also the historic one adopted since the first neuromorphic chips from the late 1980s. It takes its roots in neuroscience observations and then attempts at (i)~replicating these observations \textit{in silico}, and (ii) integrating them at scales ranging from hundreds or thousands~\cite{Seo11,Park14,Qiao15,Mayr16,Moradi18,Frenkel19a,Stuijt21,Frenkel19b} to millions of neurons~\cite{Schemmel10,Painkras13,Benjamin14,Akopyan15,Davies18}, leading to a tradeoff between \textit{versatility} and \textit{efficiency}. Integrations reaching a billion neurons can be achieved when racks of neuromorphic chips are assembled in a supercomputer setup. The simulation in real time of about 1\% of the human brain is currently possible~\cite{Furber14}, and of the full human brain within a few years~\cite{Mayr19}. Bottom-up approaches thus allow designing experimentation platforms that support acceleration of neuroscience simulations~\cite{Schemmel10}, brain reverse-engineering through \textit{analysis by synthesis}~\cite{Cauwenberghs13,Indiveri15a} and even the exploration of hybrid setups between biological and artificial neurons~\cite{Vogelstein08,George15}. Their application to brain-machine interfaces~\cite{Corradi15,Boi16} and closed sensorimotor loops for autonomous cognitive agents~\cite{Sandamirskaya14,Conradt15,Milde17,Indiveri19} is also under investigation. However, the inherent difficulty of bottom-up approaches lies in applying the resulting hardware to real-world problems beyond the scope of neuroscience-oriented applications, a point that is further emphasized by the current lack of appropriate and widely-accepted neuromorphic benchmarks~\cite{Davies19}. Therefore, bottom-up designs have so far been mostly used for basic research. In this section, as highlighted in Fig.~\ref{fig_strategy}, we follow the steps of the bottom-up approach by surveying neuromorphic designs from the building block level (Section~\ref{ssec_bb}) to their silicon integration (Section~\ref{ssec_siep}).

\begin{figure}
\centering
\noindent\includegraphics[width=1.0\columnwidth]{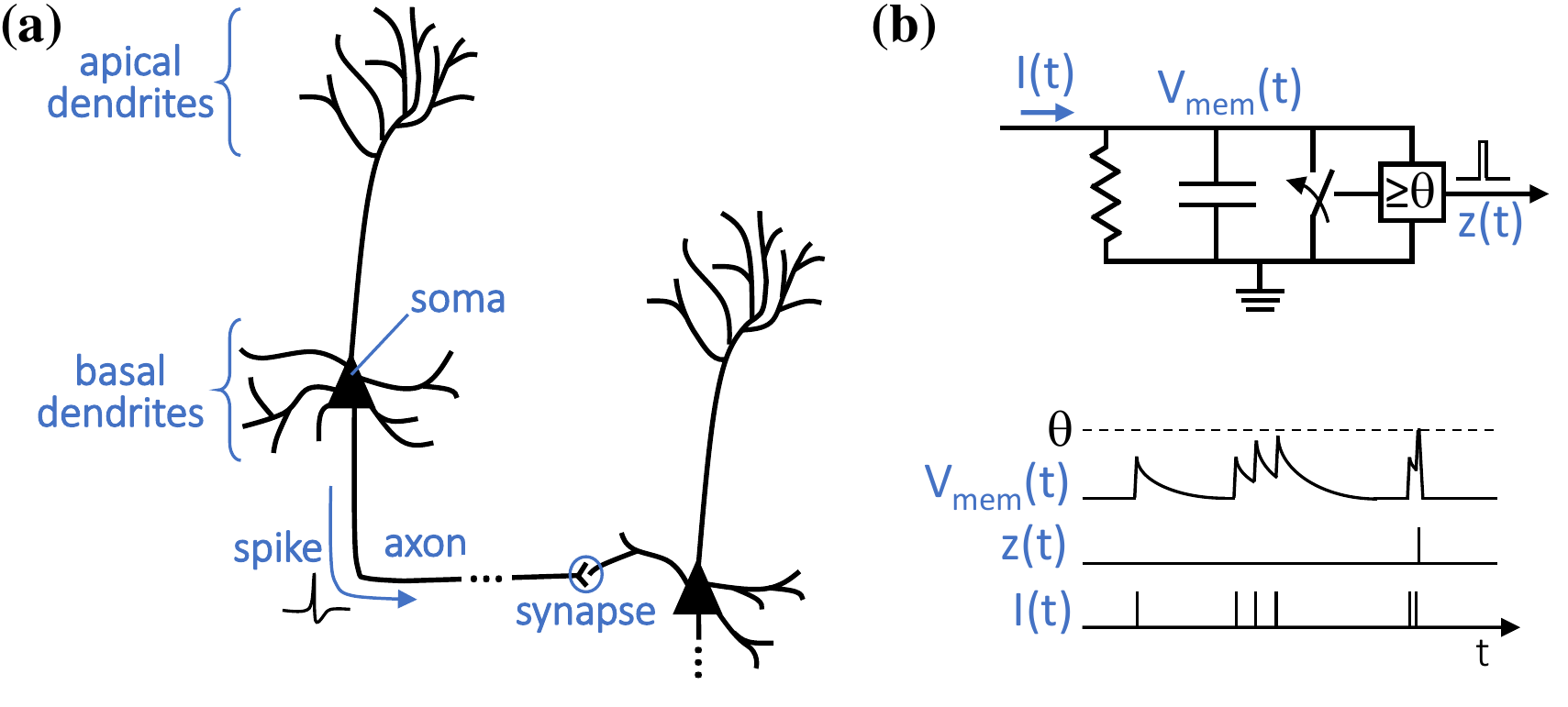}
\caption{Simplified neuron morphology and modeling. \textbf{(a)} Neurons are composed of a soma, an axon, and dendrites (in pyramidal neurons, apical dendrites receive feedback from higher-order brain areas, basal dendrites are close to the soma and receive feedforward sensory inputs). Adapted and extended from~\cite{Gerstner14}. \textbf{(b)}  The leaky integrate-and-fire (LIF) neuron model is a first-order approximation of the biological neuron as an RC filter with a spiking non-linearity and a reset mechanism. The firing threshold is denoted as $\theta$, the membrane potential as $V_\text{mem}(t)$, the input dendritic current as $I(t)$, and the spiking output as $z(t)$.}
\label{fig_introNeurons}
\end{figure}

\subsection{Building blocks}\label{ssec_bb}

As the key computational elements of biological systems, the \textit{neurons} carry out nonlinear transformations of their inputs, both in space and time, and are divided into three stages (Fig.~\ref{fig_introNeurons}): the \textit{dendrites} act as an input stage, the core computation takes place in the \textit{soma} and the outputs are transmitted along the \textit{axon}, which connects to dendrites of other neurons through \textit{synapses}. The soma, often simply referred to as a neuron in neuromorphic systems, is covered in Section~\ref{sssec_neurons}. The synapses, dendrites and axons are then covered in Sections~\ref{sssec_synapses}, \ref{sssec_dendrites} and~\ref{sssec_axons}, respectively. The neural tissue also contains glial cells, which are believed to take a structuring and stabilizing role~\cite{Jessen04} with a few silicon implementations~\cite{Nazari15,Irizarry-Valle15}, but whose study is beyond the scope of this survey.

\vspace*{3mm}\subsubsection{Neurons (soma)}\label{sssec_neurons}~\\\vspace*{-3.5mm}

One of the simplest neuron models, which originates from the work of Louis Lapicque in 1907~\cite{Lapicque07}, describes biological neurons as \textit{integrating} synaptic currents into a membrane potential and \textit{firing} a spike (i.e.~action potential) when the membrane potential exceeds a firing threshold, after which the membrane potential is reset. It is thus referred to as the \textit{integrate-and-fire} (I\&F) model, while the addition of a leakage term leads to the \textit{leaky integrate-and-fire} (LIF) model, which emphasizes the influence of recent inputs over past activity~\cite{Burkitt06}. This basic linear-filter operation can be modeled by an RC circuit. The widespread I\&F and LIF models are \textit{phenomenological} models: they aim at computational efficiency while exhibiting, from an input/output point of view, a restricted repertoire of biophysical behaviors chosen for their prevalence or relevance for a specific application. On the other end of the neuron models spectrum, \textit{conductance-based} models aim at a faithful correspondence with the biophysics of biological neurons. The Hodgkin-Huxley~(H\&H) model~\cite{HodgkinHuxley52} lies the closest to the biophysics but is computationally-intensive as it consists of four nonlinear ordinary differential equations. The Izhikevich model is a two-dimensional reduction of the H\&H model~\cite{Izhikevich03} that can still capture the 20 main behaviors of biological spiking neurons found in the cortex~\cite{Izhikevich04}, but whose parameters have lost correspondence with the biophysics. The adaptive-exponential~(AdExp) two-dimensional model is similar to the Izhikevich model and differs by the non-linearity in the spiking mechanism, which is exponential instead of quadratic~\cite{Brette05}. Due to this exponential, the AdExp neuron model suits well a subthreshold analog design approach and can be seen as a generalized form of the Izhikevich model. We refer the reader to~\cite{Izhikevich04} for a detailed neuron model summary.

\begin{figure}
\centering
\noindent\includegraphics[width=0.87\columnwidth]{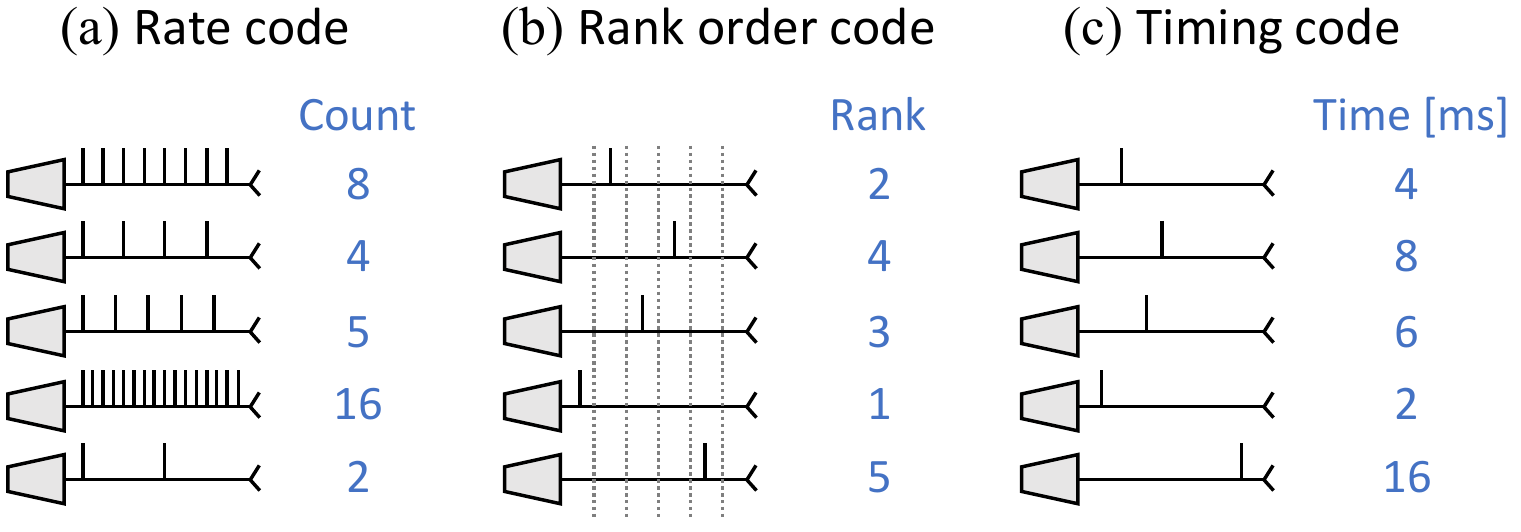}
\caption{Main encodings in spiking neural networks, as defined in~\cite{Thorpe01}. The neuron axons represent a time axis, the most recent spikes being closest to the soma. (\textbf{a}) Conventional rate code, easy to use and accurate but inefficient in its spike use. (\textbf{b}) Rank order code, efficient in its spike use but with a limited representational power. (\textbf{c}) Timing code in the specific case of time-to-first-spike (TTFS) encoding, both efficient in its spike use and accurate, illustrated for an arbitrary resolution of 1\,ms.}
\label{fig_codes}
\end{figure}

The choice of the neuron model is also intrinsically tied to the target neural coding approach. As the I\&F neuron model only behaves as an integrator, it does not allow leveraging complex temporal information~\cite{Indiveri10}. Therefore, the I\&F model is usually restricted to the use of the \textit{rate code} (Fig.~\ref{fig_codes}(a)), a standard spike coding approach directly mapping continuous values into spike rates~\cite{Thorpe01}. It is a popular code due to its simplicity, which also allows for straightforward mappings from ANNs to SNNs~\cite{Diehl15,Diehl16,Rueckauer17}, at the expense of a high power penalty as each spike only encodes a marginal amount of information. This aspect can be partly mitigated with the use of the \textit{rank order code} (Fig.~\ref{fig_codes}(b)), sometimes used as an early-stopping variant of the rate code, without taking into account relative timings between spikes. Behavior versatility is thus necessary to explore codes that embed higher amounts of data bits per spike and favor sparsity by leveraging time, such as the \textit{timing code}~\cite{Thorpe01,Arthur06,Yu13,Frenkel21}, where the popular \textit{time-to-first-spike} (TTFS) variant encodes information in the time taken by a neuron to fire its first spike (Fig.~\ref{fig_codes}(c)). In order to efficiently exploit temporal codes, neurons must capture time into computation~\cite{Izhikevich04}. We discuss in~\cite{Frenkel19a} how the 20 Izhikevich behaviors of biological cortical spiking neurons offer a variety of ways to do so.

\begin{figure}[!t]
\centering
\noindent\includegraphics[width=0.99\columnwidth]{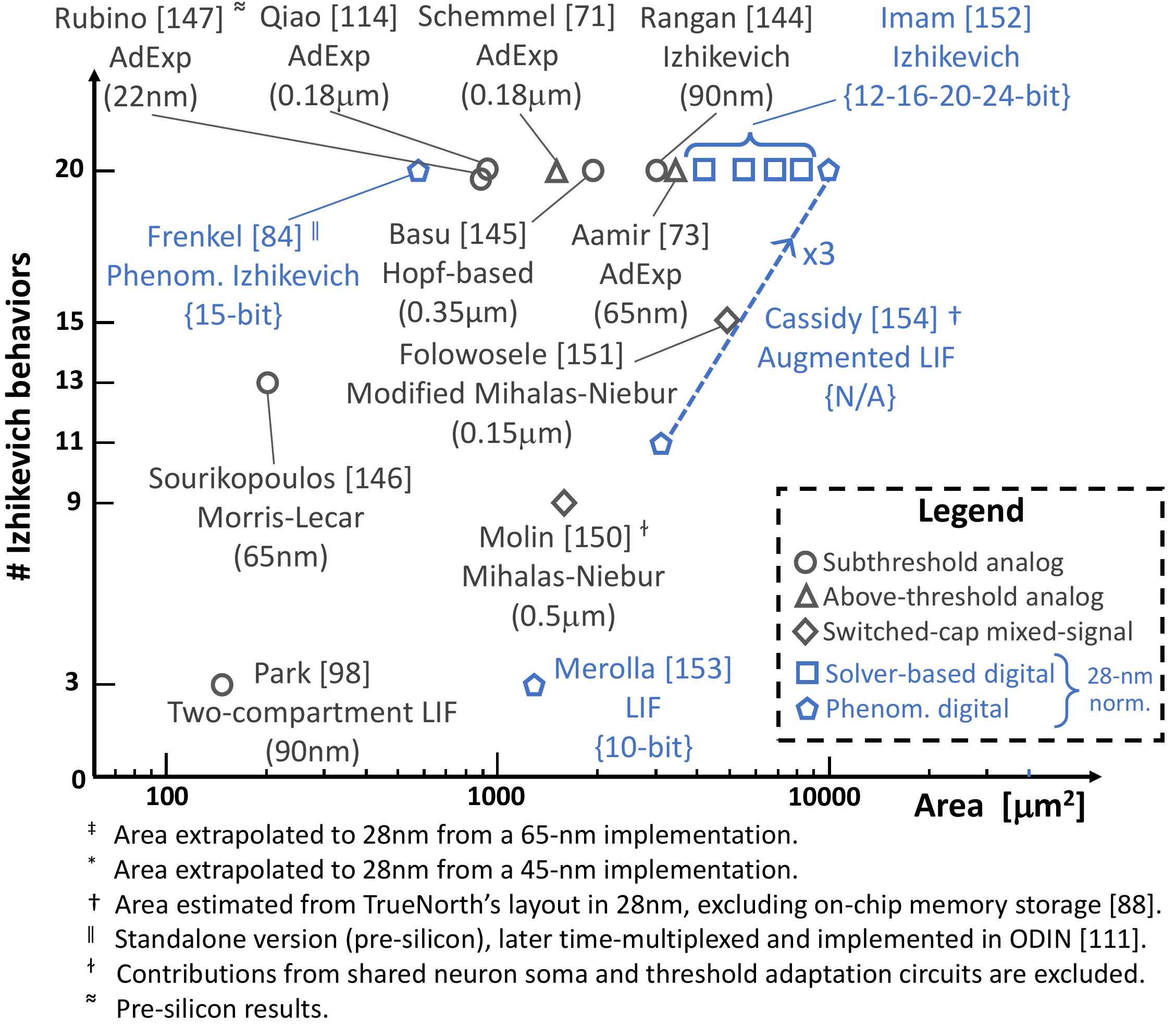}
\caption{State of the art of analog and digital neuron implementations: versatility (measured in the number of Izhikevich behaviors) against area tradeoff. The area of digital designs has been normalized to a 28-nm node using the node factor. This normalization has not been applied to analog designs as they require redesign to compensate for performance degradation during technology scaling: original area and technology node are reported. All neurons presented in this figure are standalone (i.e., non time-multiplexed), except in~\cite{Cassidy13a} for which only the update logic area is reported, and in~\cite{Molin17} for which contributions from shared soma and threshold adaptation circuits are excluded. The designs from~\cite{Schemmel10,Qiao15,Aamir18a,Rubino21} emulate an adaptive-exponential neuron model and are thus believed to reach the 20 Izhikevich behaviors~\cite{Naud08}, though not demonstrated. Adapted and extended from~\cite{Frenkel17a}.\vspace*{1.5mm}}%
\label{fig_neurons}
\end{figure}

Therefore, the tradeoff between biophysical versatility and implementation efficiency of silicon neurons is strongly dependent on the underlying model, the target code, and whether an emulation or a simulation implementation strategy is pursued (Table~\ref{tab_design}). An overview of the current state of the art for analog, mixed-signal, and digital neurons is provided in Fig.~\ref{fig_neurons}. Only standalone non-time-multiplexed neuron implementations are shown for a fair comparison of their versatility/efficiency tradeoff, measured here by the number of Izhikevich behaviors and the silicon area, respectively. The physics-based emulation approach pursued with subthreshold analog design achieves overall excellent versatility/efficiency tradeoffs~\cite{Rangan10,Basu10,Qiao15,Sourikopoulos17,Rubino21}, followed closely by model-based above-threshold analog designs~\cite{Schemmel10,Aamir18a}. By their similarity with the Izhikevich model, which is implemented in~\cite{Rangan10}, AdExp neurons are believed to reach the 20 Izhikevich behaviors~\cite{Naud08}, although it has not been demonstrated in their silicon implementations in~\cite{Schemmel10,Aamir18a,Qiao15,Rubino21}. The conductance- and Hopf-bifurcation-based neuron of~\cite{Basu10} is also able to reproduce the full repertoire of Izhikevich behaviors. Neuron implementations from~\cite{Benjamin14} and~\cite{Wijekoon08} should provide similar tradeoffs, but no information is provided as to their number of Izhikevich behaviors. With a reduced number of behaviors, mixed-signal SC implementations of the Mihalas-Niebur model in~\cite{Molin17} and~\cite{Folowosele09} were demonstrated to exhibit 9 and 15 out of the 20 Izhikevich behaviors, respectively, although with relatively high area due to their older technology node. The Morris-Lecar model is also explored in~\cite{Sourikopoulos17} and is believed to reach 13 out of the 20 Izhikevich behaviors~\cite{Izhikevich04}. The phenomenological approach is followed in~\cite{Park14} with LIF neurons in an extended two-compartment version that models separate dendritic voltages. On the other hand, digital designs release the constraints on design time and sensitivity to noise, mismatch and PVT variations at the expense of going for a simulation approach lying further from the biophysics, thus inducing overall a large area penalty compared to analog designs. This is illustrated in the neuron implementation from~\cite{Imam13} that implements a timestepped solver for the differential equations of the Izhikevich neuron model, while the phenomenological approach is followed in~\cite{Merolla11} with a 10-bit LIF neuron. Between both approaches lies the neuron model of Cassidy~\textit{et~al.}~\cite{Cassidy13a}, it is based on a LIF neuron model to which configurability and stochasticity are added. This model is used in the TrueNorth chip~\cite{Akopyan15} and exhibits 11 Izhikevich behaviors, while the 20 behaviors can be reached by coupling three neurons together, showing a configurable versatility/efficiency tradeoff. Finally, the event-driven phenomenological Izhikevich neuron proposed in~\cite{Frenkel17a} alleviates the efficiency gap of digital approaches by pursuing a direct implementation of the Izhikevich behaviors, not of the underlying mathematical model~\cite{Izhikevich03}.

\vspace*{3mm}\subsubsection{Synapses}\label{sssec_synapses}~\\\vspace*{-3mm}

Biological synapses embed the functions of memory and plasticity in extremely dense elements~\cite{Indiveri15a}, allowing neurons to connect with 100 to 10k incoming synapses per neuron (i.e.~fan-in)~\cite{Koch99}. Optimizing the versatility/efficiency tradeoff appears as especially critical for the synapses, as they often dominate the area of neuromorphic processors, sometimes by more than one order of magnitude~\cite{Qiao15}. In order to achieve large-scale integrations, designers often either move synaptic resources off-chip~(e.g.,~\cite{Painkras13,Benjamin14}), which comes at the expense of an increase in the system power and latency~\cite{Horowitz14}, or drop the key feature of \textit{synaptic plasticity}, thereby relying on static synaptic weights that are frozen once initialized~(e.g.,~\cite{Akopyan15,Moradi18}). However, retaining embedded online learning is important for three reasons. First, it allows low-power autonomous agents to collect knowledge and adapt to new features in uncontrolled environments, where new training data is presented on-the-fly in real time~\cite{Sandin14,Indiveri19}. Second, from a computational efficiency point of view, neuromorphic designs deprived from synaptic plasticity rely on off-chip optimizers, thus precluding deployment in applications that are power- and resource-constrained not only in the inference phase, but also in the training phase. Finally, exploring biophysically-realistic silicon synapses embedding spike-based plasticity mechanisms may help unveil how they operate in the brain and support cognition~\cite{Indiveri09}. This bottom-up analysis-by-synthesis step (Fig.~\ref{fig_strategy}) may also ideally complement top-down research in bio-plausible error backpropagation algorithms (see Section~\ref{ssec_algos}). Therefore, a careful hardware-aware selection of spike-based synaptic plasticity rules is necessary for the design of efficient silicon synapses.

A wide range of plasticity mechanisms is believed to take place at different timescales in the brain, where it is common to segment them into four types~\cite{Azghadi14,Indiveri15a,Zenke15,Indiveri15c}, listed hereafter starting with the shortest timescales. First, \textit{short-term plasticity} (STP) operates over milliseconds and covers short-term synaptic adaptation mechanisms such as short-term facilitation (STF) and short-term depression (STD), which have useful properties for efficient coding and multiplexing of spiking signals~\cite{Zucker02,Friauf15,Payeur21}. A few analog CMOS implementations of STP have been proposed, e.g.~in~\cite{Qiao15,Mayr16}. Second, \textit{long-term plasticity} mechanisms operate over tens to hundreds of milliseconds and cover spike-based plasticity rules, as well as working memory dynamics~\cite{Amit92}. Third, \textit{homeostatic plasticity} operates over tens to hundreds of seconds and allows scaling synaptic weights to stabilize the neuron firing frequency ranges, and thus the network activity~\cite{Turrigiano04}. There is a particular interest for homeostatic plasticity in analog designs so as to compensate for PVT variations at the network level~\cite{Bartolozzi08}. The design of efficient strategies for circuit implementations of homeostaticity is not yet mature: achieving long homeostatic timescales in analog CMOS design is challenging, although solutions have been proposed for subthreshold design in~\cite{Qiao17}, while it incurs high control and memory access overheads in time-multiplexed digital designs. Finally, \textit{structural plasticity} operates over days to modify the network connectivity~\cite{Lamprecht04}. It is usually applied to the mapping tables governing system-level digital spike routers (see Section~\ref{sssec_axons}).%

\begin{figure}
\centering
\noindent\includegraphics[width=\columnwidth]{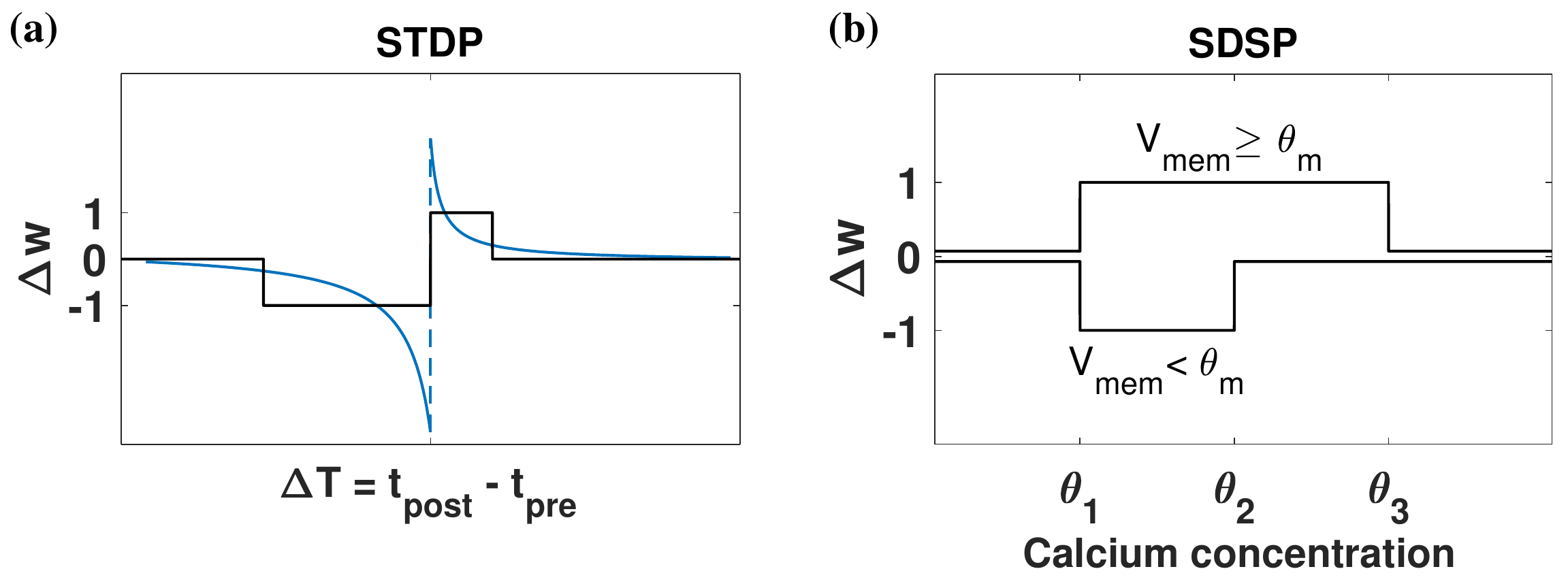}
\vspace*{-6mm}%
\caption{Illustration of the STDP and SDSP spike-based learning rules. In order to highlight their suitability for digital design, the amplitude scaling factors of SDSP and of the digital version of STDP have been normalized for unit weight updates $\Delta w$. (\textbf{a}) STDP learning rule (blue) with the popular approximation proposed by Cassidy~\textit{et~al.}~in~\cite{Cassidy11} (black). (\textbf{b}) SDSP learning rule from Brader~\textit{et~al.}~\cite{Brader07}. Adapted from~\cite{Frenkel17b,Frenkel19a}.}
\label{fig_rules}
\end{figure}

As the timescale of long-term plasticity rules is usually appropriate to perform training on spike-based image and sound classification tasks, an important body of work covers their silicon implementations, whose implementations in the mixed-signal domain have recently been reviewed in~\cite{Khacef22}. Being one of the first formulations of a long-term spike-based plasticity mechanism relying on experimental data derived by Bi and Poo~\cite{Bi98}, pair-based spike-timing-dependent plasticity~(STDP) is a conceptually simple and popular learning rule for silicon synapses~\cite{Schemmel06,Arthur06,Tanaka09,Seo11,Ramakrishnan11,CruzAlbrecht12,Bamford12}. STDP is a two-factor Hebbian learning rule relying on the relative timing of pre- and post-synaptic spikes occurring at times $t_\text{pre}$ and $t_\text{post}$, respectively. STDP strengthens correlation in the pre- and post-synaptic activities by increasing (resp.~decreasing) the synaptic weight for causal (resp.~anti-causal) orderings between pre- and post-synaptic spikes. It follows an exponential shape shown as a blue line in Fig.~\ref{fig_rules}(a). A phenomenological implementation is proposed by Cassidy~\textit{et~al.} in~\cite{Cassidy11} for digital implementations and is shown in black in Fig.~\ref{fig_rules}(a). The STDP learning rule has been declined in various shapes and has been formulated either based on spike times or on spike order (see~\cite{Yousefzadeh18} for a recent overview). Although a spike-order-based formulation allows reducing hardware requirements by eliminating the need for precise spike times~\cite{Roclin13,Yousefzadeh18}, it does not solve the main hardware efficiency issue of the STDP rule: its non-locality in time. Indeed, computing spike time differences will always imply buffering overhead.

The spike-driven synaptic plasticity~(SDSP) learning rule proposed by Brader~\textit{et~al.}~in~\cite{Brader07} led to several silicon implementations~\cite{Giulioni08,Mitra09,Qiao15,Mayr16,Frenkel17b,Frenkel19a,Frenkel19b}. Instead of relying on relative pre- and post-synaptic spike timings, SDSP computes synaptic weight updates based on the internal state of the post-synaptic neuron at the time of the pre-synaptic spike, thereby leading to a learning rule that is local in both space and time. If the post-synaptic membrane voltage $V_{\text{mem}}$ is above (resp.~below) a given threshold $\theta_m$, the synaptic weight undergoes a step increase (resp.~decrease) upon the arrival of a pre-synaptic spike (Fig.~\ref{fig_rules}(b)). Similarly to STDP, SDSP strengthens correlation between pre- and post-synaptic activities as the membrane potential indicates whether or not the post-synaptic neuron is about to spike. In order to improve the recognition of highly-correlated patterns, Brader~\textit{et~al.}~add a stop-learning mechanism based on the Calcium concentration of the post-synaptic neuron~\cite{Brader07}. The Calcium concentration provides an image of the recent post-synaptic firing activity: if it is beyond average ranges (thresholds $\theta_1$, $\theta_2$ and $\theta_3$ in Fig.~\ref{fig_rules}(b)), there is evidence that learning already took place and that further potentiation or depression is likely to result in overfitting. The learning ability of SDSP is similar to that of STDP but presents better biophysical accuracy and generalization properties~\cite{Brader07}. Both STDP and SDSP require careful hyperparameter tuning to achieve acceptable performance levels~\cite{Frenkel19a,Frenkel19b,Safa22a}.

Overall, the specific learning rule and resolution selected for the design determines the synapse circuit size, the learning ability, and the memory lifetime of the network as a function of the number of new stimuli received~(i.e.~how long a learned pattern can be reliably retrieved as synaptic weights adapt, also known as \textit{the palimpsest property})~\cite{Fusi02}. A particularly important aspect for the choice of the spike-based learning rule is its impact on the memory architecture, which will in turn define how tightly memory and computation can be co-integrated (see Section~\ref{ssec_design_tmux}). In particular, current high-density integrations with on-chip synaptic weight storage usually rely on SRAM~(see Section~\ref{ssec_siep}). Indeed, standard single-port foundry SRAMs currently have densities as high as 0.120$\mu$m$^2$/bit in 28-nm FDSOI CMOS~\cite{Thomas14} or 0.031$\mu$m$^2$/bit in the recent Intel 10-nm FinFET node~\cite{Mistry17}. Foundry SRAMs are thus an efficient substrate for low-cost synapse array design, which suits well a time-multiplexed approach. However, the memory access patterns required by the considered learning rule might imply the use of custom SRAMs instead of single-port foundry SRAMs, thus automatically inducing design time and density penalties as the layout design rule checks (DRC) for logic must be used instead of the foundry bitcell pushed rules~\cite{Chen11}. This is a known issue for spike-timing-based rules as their non-locality in time implies complex memory access patterns (e.g., see~\cite{Seo11}, where a custom dual-port SRAM with both row and column accesses was designed), while SDSP-derived rules were shown to be compatible with single-port foundry SRAMs as they only rely on information available~\mbox{locally in both space and time~\cite{Frenkel17b,Frenkel19a,Frenkel19b}.}

However, purely local learning rules relying on local pre- and post-synaptic activities (i.e.~\textit{two-factor} rules) are unable to accommodate for dependence on higher-order feedback: adding a third modulation factor is necessary to represent global information (output-prediction agreement, reward, surprise, novelty or teaching signal), and to relate it to local input and output activities for \textit{synaptic credit assignment}~\cite{Roelfsema18}, thereby leading to \textit{three-factor} rules. Just as the Calcium concentration in SDSP corresponds to a third factor modulating the pre- and post-synaptic activities, several other third-factor learning rules have been proposed, including the Bienenstock-Cooper-Munro (BCM) model~\cite{Bienenstock82}, the triplet-based STDP~\cite{Pfister06}, and several other variants of STDP and SDSP, e.g.~\cite{Graupner10,Urbanczik14}, from which the silicon synapse design of~\cite{MH16} is inspired. Furthermore, as the global modulation signal may be delayed over second-long behavioral timescales, there is a need for synapses to maintain a memory of their past activity, which may be achieved through local synaptic \textit{eligibility traces}~\cite{Gerstner18}. While the computation of eligibility traces is already supported by some neuromorphic platforms with the help of von Neumann co-processors~\cite{Painkras13,Davies18,Grubl20}, a time-multiplexed digital implementation was recently demonstrated in~\cite{Frenkel22}. A fully-parallel implementation was also proposed in~\cite{Demirag21} by exploiting the conductance drift non-ideality of phase change memory (PCM) devices. This growing complexity in synaptic learning rules is closely related to dendritic computation (Section~\ref{sssec_dendrites}).

\subsubsection{Dendrites}\label{sssec_dendrites}~\\\vspace*{-3.5mm}

While the theory of synaptic plasticity focused first on point spiking neuron models (i.e.~single-compartment neurons consisting only of the soma and the synapses, without dendrites, as defined in Fig.~\ref{fig_introNeurons}) and two-factor learning rules driven by the correlation between the pre- and post-synaptic spike timings, it now appears that STDP-based learning rules emerge as a special case of a more general plasticity framework~\cite{Bengio17,Ebner19}. Although not fully characterized yet, several important milestones toward this general plasticity framework appear to involve dendritic functions. First, correlating pre-synaptic spikes with the post-synaptic membrane voltage and its low-pass-filtered version, which could correspond to a local dendritic voltage, allows accommodating for most experimental effects that cannot be explained by STDP alone~\cite{Clopath10}. Second, the local dendritic potentials in multi-compartment neuron models are shown to predict activity in the soma (i.e.~predictive coding), with implications in supervised, unsupervised and reinforcement learning setups~\cite{Urbanczik14}. Finally, combining a detailed dendritic model of a cortical pyramidal neuron with a single general plasticity rule strongly grounded on the biophysics (i.e. local low-pass-filtered voltage traces at the pre- and post-synaptic sites) could unify previous theoretical models and experimental findings~\cite{Ebner19}. Therefore, dendrites emerge as a key ingredient that allows generalizing STDP, providing a neuron-specific feedback and potentially enabling error-based synaptic credit assignment in the brain. Furthermore, new top-down algorithms mapping onto dendritic primitives also give a strong incentive for neuromorphic hardware supporting dendritic processing (see Section~\ref{ssec_algos}). For these reasons, although only a few earlier works investigated the design of dendritic circuits~\cite{Hasler07,Parker08,Wang10,Hsu14}, silicon implementations of dendrites and multi-compartment neuron models are now receiving an increasing interest~\cite{Schemmel17,Davies18,Benjamin21,Rubino22}.

\vspace*{3mm}\subsubsection{Axons}\label{sssec_axons}~\\\vspace*{-3.5mm}

Neurons communicate spikes through their axon, which covers both short- and long-range connectivity. While the neuron and synapse implementation can be analog, mixed-signal or digital, the spike distribution infrastructure is always implemented digitally to allow for a high-speed communication of spike events on shared bus resources with a minimized footprint~\cite{Liu14}. The standard protocol for spike communication is the asynchronous address-event representation (AER)~\cite{Mortara94,Boahen00}, from simple point-to-point links in small-scale designs~\cite{Qiao15,Mayr16,Frenkel19a} to complex network-on-chip (NoC) infrastructures allowing for large-scale on- and off-chip integration~\cite{Navaridas09,Schemmel10,Benjamin14,Akopyan15,Park17,Moradi18,Davies18,Frenkel19b}. While point-to-point links cannot scale efficiently as they require the use of dedicated external routers, large-scale infrastructures ensure that several chips can be interconnected directly through their on-chip routers. We refer the reader to~\cite{Park17} for a review on linear, mesh-, torus- and tree-based router types.

Given constraints on the target network structure, such as the fact that biological neural networks typically follow a dense local and sparse long-range connectivity (i.e.~\textit{small-world} connectivity~\cite{Bassett06}), an efficient routing infrastructure must maximize the fan-in and fan-out connectivity while minimizing its memory footprint. Common techniques to optimize this tradeoff include a two- or three-level hierarchical combination of different router types (e.g.,~\cite{Park17,Moradi18,Frenkel19b,Leite22}), and of source- and destination-based addressing. In the former, source neurons are agnostic of the implemented connectivity, only the source neuron address is sent over the NoC. In exchange, this scheme requires routers to implement mapping tables, and thus to have access to dedicated memory resources, which can be either off-chip~\cite{Benjamin14,Park17} or on-chip~\cite{Navaridas09,Moradi18} depending on the target tradeoff between efficiency and flexibility. On the other hand, in the latter, the source neuron sends a destination-encoded packet over the NoC. This allows for low-cost high-speed memory-less routers, at the expense of moving the connectivity memory overhead at the neuron level~\cite{Akopyan15,Frenkel19b}. These different hierarchical combinations of router types and of source- and destination-based addressing allow reaching different tradeoffs between scalability, flexibility and efficiency, which will become apparent when quantitatively comparing experimentation platforms in Section~\ref{sssec_bottomup_tradeoff}.

\vspace*{3mm}\subsection{Silicon integration}\label{ssec_siep}

Based on the neuron, synapse, dendrite and axon building blocks described in Section~\ref{ssec_bb}, small- to large-scale integrations \textit{in silico} have been achieved with a wide diversity of design styles and use cases. Here, we review these designs, first qualitatively to outline their applicative landscape (Section~\ref{sssec_bottomup_overview}), then quantitatively to assess the key versatility/efficiency tradeoff that bottom-up designs aim at optimizing (Section~\ref{sssec_bottomup_tradeoff}). Finally, we highlight the challenges faced by a purely bottom-up design approach when efficient scaling to real-world tasks is required (Section~\ref{sssec_bottomup_learning}).

\vspace*{3mm}\subsubsection{Overview of neuromorphic experimentation platforms}\label{sssec_bottomup_overview}~\\\vspace*{-2mm}

Depending on their implementation and chosen circuit design styles, bottom-up neuromorphic experimentation platforms can be used as testbeds for neuroscience-oriented applications if they aim at replicating the biophysics, either through emulation or simulation of detailed models (see Section~\ref{sec_design}). Small-scale systems can also support bio-inspired edge computing applications, which will be further discussed in Section~\ref{sec_discussion}. Finally, large-scale systems usually target high-level functional abstractions of neuroscience, i.e.~\textit{cognitive computing}. In the following, we review the applicative landscape of analog and mixed-signal designs, followed by digital ones. A qualitative overview is provided in Table~\ref{table_expePlat}. 

\begin{table}[!t]
\caption{Bottom-up neuromorphic experimentation platforms overview. (S)~denotes small-scale chips embedding up to 512 neurons.\\(M)~denotes medium-scale chips embedding 1k to 2k neurons with a large-scale communication infrastructure.\\(L)~denotes large-scale chips or systems, from 10k-100k neurons (single chip/wafer) to millions of neurons (multi-chip setups), with up to a billion neurons for supercomputer setups.\vspace*{-0.7mm}}
\label{table_expePlat}
\renewcommand{\arraystretch}{0.97}
\centering
\resizebox{0.97\columnwidth}{!}{
\begin{tabular}{cccccc}
\toprule%
\multicolumn{3}{c}{Implementation} & Key designs$^\ddag$ & & Main application \\\midrule
\rule{0pt}{-2ex}\\
\multirow{10}{*}[-3mm]{\shortstack{Analog\\mixed-signal}} & \rdelim\{{11}{10pt}[] & \multirow{3}{*}{Subthreshold} & \multirow{3}{*}{\shortstack{ROLLS (S)~\cite{Qiao15}\\ DYNAPs (M)~\cite{Moradi18}\\ Neurogrid (L)~\cite{Benjamin14}}} & & \multirow{3}{*}[-1mm]{\shortstack{Brain emulation,\\ basic research and\\edge computing (S-M)}}\\ \\ \\
& & & & & \\
& & \multirow{4}{*}{Above-threshold} & \multirow{4}{*}{\shortstack{HICANN (S)~\cite{Schemmel10}\\ HICANN-X (S)~\cite{Schemmel20}\\ BrainScaleS (L)~\cite{Schemmel10}\\ (BrainScaleS 2) (L)$^*$~\cite{Pehle22}}} & & \multirow{4}{*}{\shortstack{Neuroscience\\simulation\\acceleration}}\\ \\ \\ \\
& & & & & \\
& & \multirow{2}{*}{\shortstack{Switched- or\\ time-muxed-cap}} & \multirow{2}{*}{\shortstack{\textit{Mayr et al.}~(S)~\cite{Mayr16}\\ IFAT (L)~\cite{Park14}}} & & \multirow{2}{*}[-1mm]{\shortstack{Bio-inspired edge to\\cognitive computing}}\\ \\
& & & & & \\%
& & & & & \\%
\multirow{28}{*}[-4.3mm]{Digital} & \rdelim\{{30}{10pt}[] & \multirow{8}{*}[-1.7mm]{Software-based$^\dag$} & \multirow{6}{*}{\shortstack{GENESIS~\cite{Bower98}\\ NEURON~\cite{Carnevale06} \\ NEST~\cite{Gewaltig07} \\ Auryn~\cite{Zenke14} \\ EDEN~\cite{Panagiotou22} \\ Brian 1, 2~\cite{Goodman08,Stimberg19} \\ ANNarchy~\cite{Vitay15} \\ GeNN~\cite{Yavuz16}}} & & \multirow{8}{*}[-2.5mm]{\shortstack{Low-cost and \\flexible neuro-\\science simulation}} \\ \\ \\ \\ \\ \\ \\ \\%
& & & & & \\
& & \multirow{3}{*}{\shortstack{Distributed\\von Neumann}} & \multirow{3}{*}{\shortstack{SpiNNaker (L)~\cite{Painkras13}\\ (SpiNNaker 2) (L)$^*$~\cite{Liu18}}} & & \multirow{3}{*}{\shortstack{Neuroscience\\simulation\\acceleration}} \\ \\ \\ 
& & & & & \\
& & \multirow{7}{*}{\shortstack{Full-custom}} & \multirow{4}{*}{\shortstack{\textit{Seo et al.}~(S)~\cite{Seo11}\\ ODIN (S)~\cite{Frenkel19a} \\ $\mu$Brain (S)~\cite{Stuijt21} \\ MorphIC (M)~\cite{Frenkel19b}}} &  & \multirow{4}{*}[-1.5mm]{\shortstack{Bio-inspired\\edge computing}} \\ \\ \\ \\ \\
& & & \multirow{3}{*}[-1mm]{\shortstack{TrueNorth (L)~\cite{Akopyan15}\\ Loihi (L)~\cite{Davies18}}} &  & \multirow{3}{*}[-1.5mm]{\shortstack{Cognitive\\computing}} \\ \\ \\
& & & & & \\
& & \multirow{7}{*}{\shortstack{FPGA-based}} & \multirow{7}{*}{\shortstack{$\mu$Caspian (S)~\cite{Mitchell20} \\ Minitaur (L)~\cite{Neil14} \\ \textit{Cassidy et al.}~(L)~\cite{Cassidy13b} \\ \textit{Wang et al.}~(L)~\cite{Wang18} \\ RANC (L)~\cite{Mack20} \\ \textit{Luo et al.}~(L)~\cite{Luo16} \\ \textit{Yang et al.}~(L)~\cite{Yang18}}} & & \multirow{7}{*}[-1.5mm]{\shortstack{Low-cost, flexible\vspace*{-0.4mm}\\neuroscience\\simulation and\\cognitive computing}} \\ \\ \\ \\ \\ \\
& & & & & \\
\rule{0pt}{-2ex}\\
\bottomrule
\vspace*{-3mm}
\end{tabular}}

\flushleft
\scriptsize

$^\ddag$ Not exhaustive. We refer the reader to~\cite{Basu22} for a more extensive list.

$^*$ The second-generation BrainScaleS and SpiNNaker large-scale systems are currently in development. For SpiNNaker 2, only proof-of-concept prototype chips have been reported so far, which embed 4 ARM cores out of the 152 planned. For BrainScales 2, the main chip HICANN-X is already available while the wafer-scale integration is currently in development.

$^\dag$ Software-based approaches run on CPU and/or GPU hardware. The implementation scale depends on available resources and the granularity of the biophysical modeling.
\vspace*{2mm}
\end{table}

\vspace*{3mm}\paragraph{Analog/mixed-signal designs}~\\\vspace*{-2mm}

The physics-based emulation approach based on \textit{subthreshold analog} design is pursued in three main designs, which primarily target basic research and also allow for the exploration of edge computing use cases in small- to medium-scale designs. First, the 0.18-$\mu$m ROLLS chip~\cite{Qiao15} is a neurosynaptic core that embeds 256 AdExp neurons~(Section~\ref{sssec_neurons}), 64k synapses with STP and 64k synapses with SDSP~(Section~\ref{sssec_synapses}). Second, the 0.18-$\mu$m \mbox{DYNAPs} chip~\cite{Moradi18} is a quad-core 2k-neuron 64k-synapse scale-up of ROLLS whose focus is put on the spike routing and communication infrastructure, at the expense of synaptic plasticity, which has been removed. A 28-nm version of the DYNAPs chip has been designed, which includes a plastic core embedding 64 neurons and 8k 4-bit digital STDP synapses, with preliminary results reported in~\cite{DeSalvo18}. Finally, the Neurogrid, a 1-million-neuron system based on sixteen 0.18-$\mu$m Neurocore chips, was designed in order to emulate the biophysics of cortical layers~\cite{Benjamin14}. However, large-scale integration is achieved at the expense of synaptic weight storage, which has been moved off-chip, thus inducing power and latency overheads. Importantly, by aiming at a direct reproduction of biophysical phenomena, these subthreshold analog designs mainly aim at \textit{understanding by building}.

The model-based \textit{above-threshold analog} design approach allows accelerating neuroscience simulations and is pursued in the BrainScaleS wafer-scale design. It relies on 0.18-$\mu$m HICANN chips with 512 AdExp neurons and 112k 4-bit STDP synapses integrated at a scale of 352 chips per wafer~\cite{Schemmel10}. BrainScaleS thus embeds 180k neurons and 40M synapses per wafer for large-scale simulation and exploration of cortical functions, with acceleration factors ranging from $10^3$ to $10^5$ compared to biological time. The second-generation 65-nm HICANN-X chips~\cite{Schemmel20} will be used for BrainScaleS 2, whose wafer-scale integration is still in development~\cite{Pehle22}.
HICANN-X embeds 512 AdExp neurons, 128k 16-bit synapses, a programmable plasticity processor, as well as multi-compartment neuron models for dendritic computation and structural plasticity~\cite{Friedmann17,Billaudelle20}. In contrast with subthreshold analog designs, the BrainScaleS platform aims at providing a tool for neuroscientists, and thus follows a \textit{building-to-understand} approach.\vspace*{1mm}

Approaches based on \textit{switched-capacitor} and \textit{capacitor-based time multiplexing} have been proposed in~\cite{Mayr16} and~\cite{Park14}. The 28-nm chip from Mayr~\textit{et~al.}~is an interesting attempt at leveraging technology scaling by using digital control and SRAM-based weight storage, while maintaining the higher biophysical accuracy of analog designs for synaptic plasticity through SC circuits~\cite{Mayr16}. Capacitor-based time multiplexing is used for neuron membrane potential storage. This small-scale chip embeds 64 neurons and 8k 4-bit synapses with both STP and SDSP, as per the implementation described in~\cite{Noack15}. It is thus suitable for near-sensor applications at the edge, where the power and area footprints should be minimized~\cite{Bol15,Shi16}. The 65-nm integrate-and-fire array transceiver (IFAT) chip from Park~\textit{et~al.}~relies on conductance-based neuron and synapse models with capacitor-based time multiplexing~\cite{Park14}, embedding as high as 65k two-compartment integrate-and-fire neurons per chip. However, synapses do not embed synaptic plasticity and their weights are stored off-chip. This chip is thus appropriate for large-scale cognitive computing experiments with relaxed synaptic requirements.\vspace*{1mm}

Finally, solutions based on non-volatile memory and emerging devices have been proposed. As mentioned in Section~\ref{ssec_design_tmux}, co-integration of memristors with CMOS is still at an early stage. A first proof-of-concept chip has recently been proposed in~\cite{Cai19}, though only demonstrated for very small problems (e.g., classification of 5$\times$5-pixel binary patterns). It embeds 5k memristor synapses at a density of 10\,$\mu$m$^2$ per synapse, which is an order of magnitude larger than state-of-the-art digital integrations. Successful RRAM-based implementations were later reported with 256-neuron 64k-synapse cores in \mbox{0.15-$\mu$m}~\cite{Yan19} and 0.13-$\mu$m~\cite{Wan20} technology nodes at a density of 1.6\,$\mu$m$^2$ per synapse. Although promising, significant work is still required to alleviate the aspects of synaptic resolution control, mismatch, and fabrication costs toward large-scale memristor-based neuromorphic systems, but progress in this direction is likely to benefit from the recent release of open-source RRAM-based process design kits (PDKs)~\cite{Modaresi23}. As an alternative with more mature technologies, a 0.35-$\mu$m flash-based STDP design has also been proposed in~\cite{Brink13}, but embedded flash memory is difficult to scale beyond 28-nm CMOS and requires high programming voltages.

\paragraph{Digital designs}~\\\vspace*{-2mm}

While neuromorphic engineering aims at a paradigm shift from von-Neumann-based architectures to distributed ones that co-locate processing and memory, the granularity at which this paradigm shift is achieved in digital implementations strongly varies between three main approaches: software-based, distributed von Neumann, or full-custom, from high to low processing and memory separation.

\textit{Software-based} approaches run on conventional von Neumann hardware. Dedicated spiking neural network simulators such as GENESIS~\cite{Bower98}, NEURON~\cite{Carnevale06}, NEST~\cite{Gewaltig07}, Brian~\cite{Goodman08}, Auryn~\cite{Zenke14} and EDEN~\cite{Panagiotou22} allow running experiments on conventional CPUs, while simulators such as ANNarchy~\cite{Vitay15}, GeNN~\cite{Yavuz16} and Brian 2~\cite{Stimberg19} provide GPU support. Software-based approaches provide the highest flexibility and control over the neuron and synapse models and the scale of the experiments. However, using von Neumann hardware to simulate SNNs comes at the cost of power and simulation time overheads, although recent work has demonstrated that GPUs can compare favorably to a SpiNNaker-based system for cortical-scale simulations~\cite{Knight18,Knight21}.

SpiNNaker follows a \textit{distributed von Neumann} approach. It was fabricated in a 0.13-$\mu$m CMOS technology and embeds 18 ARM968 cores per chip in a globally asynchronous locally synchronous (GALS) design for efficient handling of asynchronous spike data, spanning biological to accelerated time constants~\cite{Painkras13}. SpiNNaker has been optimized for large-scale SNN experiments while keeping a high degree of flexibility, with the current supercomputer-scale setup reaching the billion of neurons, i.e.~about 1\% of the human brain~\cite{Furber14}. The second-generation SpiNNaker system is in development. Current 28-nm prototype chips embed 4 ARM Cortex M4F cores out of the 152 per chip planned for the final 22-nm SpiNNaker 2 system~\cite{Liu18}. The objective is to simulate two orders of magnitude more neurons per chip compared to the first-generation SpiNNaker: when integrated at supercomputer scale, real-time simulations at the scale of the human brain will be within reach~\cite{Hoppner18}. Therefore, similarly to BrainScaleS, SpiNNaker also follows a \textit{building-to-understand} approach.

\begin{table*}
\vspace*{2mm}
\setlength\tabcolsep{4pt}
\caption{Comparison of specifications and measured performances across bottom-up neuromorphic chips. Extended from~\cite{Frenkel19a}.\vspace*{-2mm}}
\label{table_SoA_fullComp}
\renewcommand{\arraystretch}{1.05}
\centering
\resizebox{\textwidth}{!}{\begin{tabular}{lcccccccccccc}
\toprule%

Author & Benjamin~\cite{Benjamin14} & Qiao~\cite{Qiao15} & Moradi~\cite{Moradi18} & Schemmel~\cite{Schemmel20} & Mayr~\cite{Mayr16} & Painkras~\cite{Painkras13} & Seo~\cite{Seo11} & Frenkel~\cite{Frenkel19a} & Frenkel~\cite{Frenkel19b} & Stuijt~\cite{Stuijt21} & Akopyan~\cite{Akopyan15} & Davies~\cite{Davies18} \\
Publication & PIEEE, 2014 & Front. NS, 2015 & TBioCAS, 2017 & arXiv, 2020 & TBioCAS, 2016 & JSSC, 2013 & CICC, 2011 & TBioCAS, 2019a & TBioCAS, 2019b & Front. NS, 2021 & TCAD, 2015 & IEEE Micro, 2018 \\
Chip name & Neurogrid & ROLLS & DYNAPs & HICANN-X & -- & SpiNNaker & -- & ODIN & MorphIC & $\mu$Brain & TrueNorth & Loihi \\\midrule

\multirow{2}{*}{Implementation} & \multirow{2}{*}{\shortstack{Mixed-signal\\(subthreshold)}} & \multirow{2}{*}{\shortstack{Mixed-signal\\(subthreshold)}} & \multirow{2}{*}{\shortstack{Mixed-signal\\(subthreshold)}} & \multirow{2}{*}{\shortstack{Mixed-signal\\(above-threshold)}} & \multirow{2}{*}{\shortstack{Mixed-signal\\(SC)}} & \multirow{2}{*}{\shortstack{Digital\\(GALS)}} & \multirow{2}{*}{\shortstack{Digital\\(sync)}} & \multirow{2}{*}{\shortstack{Digital\\(sync)}} & \multirow{2}{*}{\shortstack{Digital\\(sync)}} & \multirow{2}{*}{\shortstack{Digital\\(async)}} & \multirow{2}{*}{\shortstack{Digital\\(GALS)}} & \multirow{2}{*}{\shortstack{Digital\\(async)}} \\
& \\

Technology & 0.18\,$\mu$m & 0.18\,$\mu$m & 0.18\,$\mu$m & 65\,nm & 28\,nm & 0.13\,$\mu$m & 45\,nm SOI & 28\,nm FDSOI & 65\,nm & 40\,nm & 28\,nm & 14\,nm FinFET \\\vspace*{-0.4mm}%

Cores$^\diamond$ & 16 & 1 & 4 & 1 & 1 & 18 & 1 & 1 & 4 & 1 & 4096 & 128 \\

Neurosynaptic core area [mm$^2$] & 168 & 51.4 & 7.5 & 27.9 & 0.36 & 3.75 & 0.8 & 0.086 & 0.71 & 1.42 & 0.095 & 0.4 \\

State update circuits & Fully-parallel & Fully-parallel & Fully-parallel & Fully-parallel & Time-multiplexed & Time-multiplexed & Time-multiplexed & Time-multiplexed & Time-multiplexed & Fully-parallel & Time-multiplexed & Time-multiplexed \\

Time constant & Biological & Biological & Biological & Accelerated & Bio. to accel. & Bio. to accel. & Biological & Bio. to accel. & Bio. to accel. & Bio. to accel. & Biological & Bio. to accel. \\

Routing \begin{tabular}{@{}c@{}}~flexibility\\~fan-in / fan-out\end{tabular} & \begin{tabular}{@{}c@{}}Medium\\N/A\end{tabular} & \begin{tabular}{@{}c@{}}Low\\512 / 256\end{tabular} & \begin{tabular}{@{}c@{}}Medium\\64 / 4k\end{tabular} & \begin{tabular}{@{}c@{}}Medium\\256 / N/A\end{tabular} & \begin{tabular}{@{}c@{}}Low\\128 / 64\end{tabular} & \begin{tabular}{@{}c@{}}High\\Programmable\end{tabular} & \begin{tabular}{@{}c@{}}Low\\256 / 256\end{tabular} & \begin{tabular}{@{}c@{}}Low\\256 / 256\end{tabular} & \begin{tabular}{@{}c@{}}Medium\\1k / 2k\end{tabular} & \begin{tabular}{@{}c@{}}Low\\Layer-dependent\end{tabular} & \begin{tabular}{@{}c@{}}Medium\\256 / 512\end{tabular} & \begin{tabular}{@{}c@{}}High\\Programmable\end{tabular} \\

Neurons per core & 64k & 256 & 256 & 512 & 64 & max. 1000$^\curlywedge$ & 256 & 256 & 512 & 336 & 256 & max. 1024 \\

Izhikevich behaviors$^\dag$ & N/A & (20) & (20) & (20) & 3 & Programmable & 3 & 20 & 3 & 3 & 11 (3 neur: 20) & (6) \\

Synapses per core & -- & 128k & 16k & 128k & 8k & -- & 64k & 64k & 528k & 36k & 64k & 1M to 114k (1-9 bits) \\

Synaptic storage & Off-chip & Capacitor & 12-bit (CAM) & 16-bit (SRAM) & 4-bit (SRAM) & Off-chip & 1-bit (SRAM) & 4-bit (SRAM) & 1-bit (SRAM) & 4-bit (flip-flops) & 1-bit (SRAM) & 1- to 9-bit (SRAM) \\

Embedded online learning & -- & SDSP & -- & Programmable & SDSP & Programmable & S-STDP & SDSP & S-SDSP & -- & -- & Programmable \\

Neuron core density [neur/mm$^2$]$^*$ \begin{tabular}{@{}c@{}}raw\\norm.\end{tabular} & \begin{tabular}{@{}c@{}}390\\--\end{tabular} & \begin{tabular}{@{}c@{}}5\\--\end{tabular} & \begin{tabular}{@{}c@{}}34\\--\end{tabular} & \begin{tabular}{@{}c@{}}18.4\\--\end{tabular} & \begin{tabular}{@{}c@{}}178\\--\end{tabular} & \begin{tabular}{@{}c@{}}max. 267$^\circ$\\max. 5.8k\end{tabular} & \begin{tabular}{@{}c@{}}320\\826\end{tabular} & \begin{tabular}{@{}c@{}}3.0k\\3.0k\end{tabular} & \begin{tabular}{@{}c@{}}716\\3.9k\end{tabular} & \begin{tabular}{@{}c@{}}237\\483\end{tabular} & \begin{tabular}{@{}c@{}}2.6k\\2.6k\end{tabular} & \begin{tabular}{@{}c@{}}max. 2.5k\\max. 1k\end{tabular} \\

Synapse core density [syn/mm$^2$]$^*$ \begin{tabular}{@{}c@{}}raw\\norm.\end{tabular} & -- & \begin{tabular}{@{}c@{}}2.5k\\--\end{tabular} & \begin{tabular}{@{}c@{}}2.1k\\--\end{tabular} & \begin{tabular}{@{}c@{}}4.6k\\--\end{tabular} & \begin{tabular}{@{}c@{}}22.2k\\--\end{tabular} & -- & \begin{tabular}{@{}c@{}}80k\\207k\end{tabular} & \begin{tabular}{@{}c@{}}741k\\741k\end{tabular} & \begin{tabular}{@{}c@{}}738k\\4M\end{tabular} & \begin{tabular}{@{}c@{}}25k\\52k\end{tabular} & \begin{tabular}{@{}c@{}}674k\\674k\end{tabular} & \begin{tabular}{@{}c@{}}2.5M to 282k\\1M to 113k\end{tabular} \\

Supply voltage & 3.0\,V & 1.8\,V & 1.3\,V -- 1.8\,V & 1.2\,V & 0.75\,V, 1.0\,V & 1.2\,V & 0.53V -- 1.0V & 0.55\,V -- 1.0\,V & 0.8\,V -- 1.2\,V & 1.1\,V & 0.7\,V -- 1.05\,V & 0.5\,V -- 1.25\,V \\

Energy per SOP$^\ddagger$ \begin{tabular}{@{}c@{}}raw\\norm.\end{tabular} & \begin{tabular}{@{}c@{}}(941\,pJ)$^\blacktriangle$\\--\end{tabular} & \begin{tabular}{@{}c@{}}$>$77\,fJ$^\vartriangle$\\--\end{tabular} & \begin{tabular}{@{}c@{}}134\,fJ$^\vartriangle$/30\,pJ$^\blacktriangle$ (1.3\,V)\\--\end{tabular} & \begin{tabular}{@{}c@{}}(0.78\,pJ)$^\blacktriangle$\\--\end{tabular} & \begin{tabular}{@{}c@{}}$>$850\,pJ$^\blacktriangle$\\--\end{tabular} & \begin{tabular}{@{}c@{}}$>$11.3\,nJ$^\vartriangle$/26.6\,nJ$^\blacktriangle$\\$>$2.4\,nJ$^\vartriangle$/5.7\,nJ$^\blacktriangle$\end{tabular} & N/A & \begin{tabular}{@{}c@{}}8.4\,pJ$^\vartriangle$/12.7\,pJ$^\blacktriangle$ (0.55\,V)\\8.4\,pJ$^\vartriangle$/12.7\,pJ$^\blacktriangle$\end{tabular} & \begin{tabular}{@{}c@{}}30\,pJ$^\vartriangle$/51\,pJ$^\blacktriangle$ (0.8\,V)\\12.9\,pJ$^\vartriangle$/22\,pJ$^\blacktriangle$\end{tabular} & \begin{tabular}{@{}c@{}}17.6\,pJ$^\vartriangle$/61.9\,pJ$^\blacktriangle$ (1.1\,V)\\12.3\,pJ$^\vartriangle$/43.3\,pJ$^\blacktriangle$\end{tabular} & \begin{tabular}{@{}c@{}}26\,pJ$^\blacktriangle$ (0.775\,V)\\26\,pJ$^\blacktriangle$\end{tabular} & \begin{tabular}{@{}c@{}}$>$23.6\,pJ$^\vartriangle$ (0.75\,V)\\(66.1\,pJ$^\vartriangle$)\end{tabular}\\

\bottomrule
\end{tabular}}

\vspace*{1.5mm}
\begin{adjustwidth}{0.3cm}{}
\begin{spacing}{0.5}
{\flushleft\fontsize{5.8pt}{6pt}\selectfont$^\diamond$ When chips are composed of several neurosynaptic cores, we report the density numbers associated to a single core. Care should be taken that, depending on the core definition in the different chips, routing resources might be included (all single-core designs, TrueNorth, Loihi and MorphIC) or excluded (Neurogrid, DYNAPs and SpiNNaker). As opposed to the other reported designs, we consider the full Neurogrid system, which is composed of 16 Neurocore chips, each one considered as a core; routing resources are off-chip. For DYNAPs and SpiNNaker, sharing routing overhead among cores would lead to 28-\% and 37-\% density penalties compared to the reported results, respectively. The HICANN-X chip can be considered as a core of the BrainScaleS wafer-scale system. Pad area is excluded from all reported designs.}
\end{spacing}
\vspace*{3.2mm}
\end{adjustwidth}
\begin{adjustwidth}{0.3cm}{}
\begin{spacing}{0.5}
{\flushleft\fontsize{5.8pt}{6pt}\selectfont$^\dag$ By its similarity with the Izhikevich neuron model, the AdExp neuron model is believed to reach the 20 Izhikevich behaviors~\cite{Naud08}, but it has not been demonstrated in HICANN-X, ROLLS and DYNAPs. The neuron model of TrueNorth can reach 11 behaviors per neuron and 20 by combining three neurons together~\cite{Cassidy13a}. The neuron model of Loihi is based on a LIF model to which threshold adaptation is added: the neuron should therefore reach 6 Izhikevich behaviors, although it has not been demonstrated.}
\end{spacing}
\vspace*{3.2mm}
\end{adjustwidth}
\begin{adjustwidth}{0.3cm}{}
\begin{spacing}{0.5}
{\flushleft\fontsize{5.8pt}{6pt}\selectfont$^\curlywedge$ Experiment 1 reported in Table~III from~\cite{Painkras13} is considered as a best-case neuron density: 1000 simple LIF neuron models are implemented per core, each firing at a low frequency.}
\end{spacing}
\vspace*{3.1mm}
\end{adjustwidth}
\begin{adjustwidth}{0.3cm}{}
\begin{spacing}{0.5}
{\flushleft\fontsize{5.8pt}{6pt}\selectfont$^*$ Neuron (resp.~synapse) core densities are computed by dividing the number of neurons (resp.~synapses) per neurosynaptic core by the neurosynaptic core area. Regarding the synapse core density, Neurogrid and SpiNNaker use an off-chip memory to store synaptic data. As the synapse core density cannot be extracted when off-chip resources are involved, no synapse core density values are reported for these chips. Values normalized to a 28-nm CMOS technology node are provided for digital designs using the node factor squared, at the exception of the 14-nm FinFET node of Loihi for which Intel data from~\cite{Mistry17} has been used.}
\end{spacing}
\end{adjustwidth}
\vspace*{-1.7mm}
\begin{adjustwidth}{0.3cm}{}
\begin{spacing}{0.5}
{\flushleft\fontsize{5.8pt}{6pt}\selectfont$^\ddagger$ The synaptic operation energy measurements reported for the different chips do not follow a standardized measurement process. There are two main categories for energy measurements in neuromorphic chips. On the one hand, incremental values (denoted with $^\vartriangle$) describe the amount of dynamic energy paid per each additional SOP computation, they are measured by subtracting the leakage and idle switching power consumption of the chip, although the exact power contributions taken into account in the SOP energy vary across chips. On the other hand, global values (denoted with $^\blacktriangle$) are obtained by dividing the total chip power consumption by the SOP processing rate. Values normalized to a 28-nm CMOS technology node are provided for digital designs using the node factor, including for the 14-nm FinFET node of Loihi in the absence of reliable data for power normalization in~\cite{Mistry17}. The conditions under which all of these measurements have been done can be found hereafter. For Neurogrid, a SOP energy of 941\,pJ is reported for a network of 16 Neurocore chips (1M neurons, 8B synapses, 413k spikes/s): it is a board-level measurement, no chip-level measurement is provided~\cite{Benjamin14}. For ROLLS, the measured SOP energy of 77\,fJ is reported in~\cite{Indiveri15b}, it accounts for a point-to-point synaptic input event and includes the contribution of weight adaptation and digital-to-analog conversion, it represents a lower bound as it does not account for synaptic event broadcasting. For DYNAPs, the measured SOP energy of 134\,fJ at 1.3\,V is also reported in~\cite{Indiveri15b}, while the global SOP energy of 30\,pJ can be estimated from~\cite{Moradi18} using the measured 800-$\mu$W power consumption with all 1k neurons spiking at 100\,Hz with 25\% connectivity (26.2\,MSOP/s), excluding the synaptic input currents. For HICANN-X, the global value of 0.78\,pJ/SOP at 1.2\,V is only a best-case estimate based on the minimum 200-mW power consumption of the chip and its maximum throughput (1Gevents/s or 256GSOP/s). In the chip of Mayr~\textit{et~al.}, the SOP energy of 850\,pJ represents a lower bound extracted from the chip power consumption, estimated by considering the synaptic weights at half their dynamic at maximum operating frequency. For SpiNNaker, an incremental SOP energy of 11.3\,nJ is measured in~\cite{Stromatias15}, a global SOP energy of 26.6\,nJ at the maximum SOP rate of 16.56\,MSOP/s can be estimated by taking into account the leakage and idle clock power; both values represent a lower bound as the energy cost of neuron updates is not included. For ODIN and MorphIC, both incremental and global SOP energy values are provided and include power contributions from all blocks~\cite{Frenkel19a,Frenkel19b}. The global energy per SOP is measured at the maximum acceleration factor. The global energy per SOP for ODIN in biological time is 54\,pJ. For $\mu$Brain, the reported numbers were extracted during MNIST benchmarking where static and dynamic power amount to 58\,$\mu$W and 23\,$\mu$W, respectively, with 4.2\,ms and 5500 SOPs in average per sample (private communication from the authors). For TrueNorth, the measured SOP energy of 26\,pJ at 0.775\,V is reported in~\cite{Merolla14}, it is extracted by measuring the chip power consumption when all neurons fire at 20\,Hz with 128 active synapses. For Loihi, a minimum SOP energy of 23.6\,pJ at 0.75\,V is extracted from pre-silicon SDF and SPICE simulations, in accordance with early post-silicon characterization~\cite{Davies18}; it represents a lower bound as it includes only the contribution of the synaptic operation, without taking into account the cost of neuron update and learning engine update.}
\vspace*{-1mm}
\end{spacing}
\end{adjustwidth}
\end{table*}

\textit{Full-custom} digital hardware allows for high-density and energy-efficient neuron and synapse integrations, thanks to memory being moved closer to computation compared to the two above-mentioned digital approaches. As full-custom digital designs rely on SRAM-based time multiplexing, this can be related to the efficiency improvement brought by caches in conventional von Neumann processors~\cite{Goodman83}. Full-custom designs can usually be configured to span biological to accelerated time constants. The 45-nm small-scale design from Seo~\textit{et~al.}~embeds 256 LIF neurons and 64k binary synapses based on a stochastic version of STDP (S-STDP)~\cite{Seo11}, it achieves high neuron and synapse densities compared to mixed-signal designs, despite the use of a custom SRAM~(Section~\ref{sssec_synapses}). Its scale thus makes it ideal for edge computing. In line with this small-scale edge computing use case, the ODIN chip embeds 256 neurons with the 20 Izhikevich behaviors and 64k SDSP-based 4-bit synapses in 28-nm CMOS~\cite{Frenkel19a}. The 65-nm MorphIC chip scales up the neurosynaptic core of ODIN in a quad-core design allowing for large-scale multi-chip setups with a total of 2k LIF neurons and more than 2M binary synapses with stochastic SDSP (S-SDSP) per chip~\cite{Frenkel19b}. Being based on SDSP, ODIN and MorphIC can leverage the density advantage of standard single-port foundry SRAMs to achieve record neuron and synapse densities (Section~\ref{sssec_synapses}). One notable exception to the SRAM-based time-multiplexed approaches in the digital domain is $\mu$Brain~\cite{Stuijt21}, which implements a recurrent 256-64-16 LIF-based network in a fully-parallel fashion with distributed flip-flop-based memories. Combined with asynchronous event-driven processing, $\mu$Brain tackles the von Neumann bottleneck at the highest granularity, at the expense of an increase in static power and silicon area (Table~\ref{tab_tmux}), as well as the introduction of a technology-specific delay element. Finally, cognitive computing applications require large-scale platforms, which is currently offered by the 28-nm IBM TrueNorth~\cite{Akopyan15} and the 14-nm Intel Loihi~\cite{Davies18} neuromorphic chips. On the one hand, TrueNorth is a GALS design embedding as high as 1M neurons and 256M binary non-plastic synapses per chip, where neurons rely on a custom model exhibiting 11 Izhikevich behaviors, or 20 behaviors if three neurons are combined~\cite{Cassidy13a}. On the other hand, Loihi is a fully asynchronous design embedding up to 180k neurons and 114k (9-bit) to 1M (binary) synapses per chip. Neurons rely on a LIF model with a configurable number of compartments to which several functionalities such as axonal and refractory delays, spike latency and threshold adaptation have been added. The spike-based plasticity rule used for synapses is programmable and eligibility traces are supported.

Finally, it should be noted that digital approaches also encompass FPGA designs, which trade off efficiency for a higher flexibility and a reduced deployment cost compared to full-custom designs. Although beyond the scope of this survey, a wide diversity of FPGA designs cover small- to large-scale cognitive computing~(e.g.,~\cite{Mitchell20,Cassidy13b,Neil14,Wang18,Mack20}) and neuroscience-oriented applications~(e.g.,~\cite{Luo16,Yang18}).%

\vspace*{3mm}\subsubsection{Versatility / efficiency comparative analysis}\label{sssec_bottomup_tradeoff}~\\\vspace*{-3.5mm}

A quantitative overview of state-of-the-art bottom-up neuromorphic chips is provided in Table~\ref{table_SoA_fullComp}. Mixed-signal designs with analog cores and high-speed digital periphery are grouped on the left~\cite{Schemmel20,Benjamin14,Qiao15,Moradi18,Mayr16}, digital designs are grouped on the right~\cite{Painkras13,Seo11,Akopyan15,Davies18,Frenkel19a,Frenkel19b,Stuijt21}. These key designs are analyzed in details here as they cover the landscape of neuromorphic circuit design styles and tradeoffs outlined in Section~\ref{sec_design}. We refer the reader to~\cite{Basu22} for an exhaustive list.

Regarding the neuron and synapse densities, numbers are overall quite low for mixed-signal designs relying on core sub- and above-threshold analog computation as they are mostly using low-density memories and/or older technology nodes. In this respect, the mixed-signal design of Mayr~\textit{et~al.} is able to exhibit higher densities as SC circuits easily scale to advanced technology nodes (see Section~\ref{sec_design}). However, through their ability to fully leverage technology scaling and through a straightforward implementation of time multiplexing, digital designs demonstrate the highest neuron and synapse densities. Considering technology-normalized numbers and equal synaptic resolutions, ODIN and MorphIC currently have the highest neuron and synapse densities reported to date. Indeed, the memory access patterns of on-chip SDSP-based learning allow for the use of high-density single-port foundry SRAMs. Loihi is also a high-density design given its extended feature set and network configurability. On the contrary, TrueNorth does not embed learning and has a restricted network configurability through low fan-in and fan-out values. However, to date, TrueNorth remains the largest-scale single-chip design with embedded synaptic weight storage. While digital designs overall achieve high neuron and synapse densities based on time multiplexing and simplified neuron and synapse models, this comes at the expense of precluding a fully-parallel emulation of network dynamics, with two clear exceptions. First, $\mu$Brain proposes an interesting fully-parallel simulation approach that, although not supporting continuous-time dynamics, still approximates a few key functions, such as leakage, in a timestepped fashion. Second, SpiNNaker can be programmed with conductance-based models at the expense of employing a solver-based digital approach, which updates the state of all neurons and synapses at every integration timestep based on computationally-expensive models, thereby limiting its power efficiency and its ability to maintain real-time operation for large networks.

For a fair comparison of the energy per synaptic operation (SOP), Table~\ref{table_SoA_fullComp} provides two definitions: the \textit{incremental} energy per SOP and the \textit{global} one. The former is the amount of dynamic energy paid for each SOP, while the latter corresponds to the overall chip power consumption divided by the SOP execution rate, which includes static power contributions, including leakage and idle switching power (see Table~\ref{table_SoA_fullComp} for details). On the analog side, the ROLLS and DYNAPs subthreshold analog designs have a very low incremental energy per SOP on the order of 100\,fJ. However, when taking the chip static energy into account, the global energy per SOP in DYNAPs increases by two orders of magnitude, which can be explained by two factors. First, fully-parallel implementations have a penalty in static power (Table~\ref{tab_tmux}). Second, the energy cost of the digital routing infrastructure of DYNAPs suffers from an implementation in an older 0.18-$\mu$m technology node. Preliminary results from a 28-nm implementation of DYNAPs show a promising global energy per SOP of 2.8\,pJ~\cite{DeSalvo18}. On the digital side, the full flexibility in neuron and synapse models offered by the SpiNNaker platform leads to a global energy per SOP on the order of tens of nJ (a few nJ if normalized to a 28-nm node). This can be partly mitigated with advanced power reduction techniques and increased hardware acceleration, which is currently being investigated for the second generation of SpiNNaker~(e.g.,~see~\cite{Hoppner17,Partzsch17,Liu18}). Full-custom digital designs have incremental and global energies per SOP on the order of a few to tens of pJ. As digital designs usually allow spanning biological to accelerated time constants, an important aspect to consider is the time constant used for the characterization of the global SOP energy, as accelerated time constants allow better amortizing the contribution from static power. For example, the 26-pJ global energy per SOP reported for TrueNorth was measured in biological time~\cite{Merolla14}, while for ODIN, the reported 12.7\,pJ/SOP was measured in maximum acceleration (this number increases to 54\,pJ in biological time, with all neurons firing at 10\,Hz)~\cite{Frenkel19a}.

Overall, Table~\ref{table_SoA_fullComp} allows clarifying the different versatility/efficiency tradeoff optimizations achieved in bottom-up neuromorphic experimentation platforms. Analog designs focus on optimizing the versatility at the level of neuronal and synaptic dynamics while maintaining power efficiency, at the expense of area efficiency. On the contrary, in digital designs, versatility cannot be obtained through fully-parallel real-time conductance-based neuronal and synaptic dynamics. Instead, it can be obtained either from a phenomenological viewpoint or at the system level, while allowing for a joint optimization with power and area efficiencies. This flexibility in optimizing between versatility and efficiency in digital designs is highlighted with platforms going from versatility-driven (e.g.,~SpiNNaker) to efficiency-driven (e.g.,~ODIN and MorphIC), through platforms aiming at a balanced tradeoff on both sides (e.g.,~Loihi). This balanced tradeoff of Loihi should be further improved with Loihi 2, which embeds key new features such as neuron model programmability, advanced memory compression and partitioning schemes, and an extended support for generalized three-factor learning rules. These advanced features are embedded while further improving density thanks to technology scaling with the latest Intel~4 node. A technology brief supported by pre-silicon results is available in~\cite{Intel21}. Finally, mixed-signal designs based on SC circuits provide an interesting middle ground by maintaining rich dynamics, while partly alleviating the density penalty of analog designs. However, a competitive energy efficiency remains to be~\mbox{demonstrated in SC neuromorphic designs.}

\vspace*{3mm}
\subsubsection{Spike-based online learning performance assessment}\label{sssec_bottomup_learning}~\\\vspace*{-3.5mm}

\begin{table}
\caption{Benchmark summary for silicon implementations of STDP- and SDSP-based learning rules. Adapted from~\cite{Frenkel19b}.}
\vspace*{-1mm}
\label{tab_bench}
\centering
\resizebox{0.98\columnwidth}{!}{
\begin{tabular}{lccc}
\toprule%
Chip(s) & Implementation & Learning rule & Benchmark \\
\midrule%
BrainScaleS~\cite{Schemmel10} & Mixed-signal & 4-bit STDP & -- \\
DYNAPs + ROLLS~\cite{Indiveri15b} & Mixed-signal & Fixed + SDSP & 8-pattern classification \\
Mayr \textit{et al.}~\cite{Mayr16} & Mixed-signal & 4-bit SDSP & -- \\
Seo \textit{et al.}~\cite{Seo11} & Digital & 1-bit S-STDP & 2-pattern recall \\
Chen \textit{et al.}~\cite{Chen18} & Digital & 7-bit STDP & Denoising / Pre-processed MNIST \\
Loihi~\cite{Davies18} & Digital & STDP-based & Pre-processed MNIST \\
ODIN~\cite{Frenkel19a} & Digital & 3-bit SDSP & 16$\times$16 deskewed MNIST \\
MorphIC~\cite{Frenkel19b} & Digital & 1-bit S-SDSP & 8-pattern classification \\
\bottomrule%
\end{tabular}}%
\end{table}

While bottom-up experimentation platforms offer efficient implementations of bio-inspired primitives, exploiting them on complex real-world tasks can be difficult. This challenge is particularly apparent for bio-plausible synaptic plasticity, as shown in  Table~\ref{tab_bench}. Indeed, to the best of our knowledge, no silicon implementation of an STDP- or an SDSP-based learning rule has so far been demonstrated on at least the full MNIST dataset~\cite{LeCun98} without any pre-processing step. Furthermore, in all cases, these learning rules are only applied to single-layer networks or to the output layer of a network with frozen hidden layers (i.e.~shallow learning). Recent studies have demonstrated STDP-based multi-layer learning in simulation~\cite{Zheng18,Tavanei19}, including for continual-learning setups~\cite{Safa22b}, but they have not yet been ported to silicon.

Another important aspect lies in weight quantization, which is commonly applied to synapses in order to reduce their memory footprint. While standard quantization-aware training techniques need to maintain a full-resolution copy of the weights to accommodate for red{fine-grained} updates (Section~\ref{ssec_algos}), neuromorphic hardware needs to carry out learning on weights that have a limited resolution not only during inference, but also during training~\cite{Frenkel19b}. This issue, combined with simple bottom-up learning rules, tends to reduce the ability of the network to discriminate highly-correlated patterns, as highlighted by the binary-weight S-STDP study in~\cite{Yousefzadeh18}. This is another reason why simple datasets with reduced overlap are selected for benchmarking, as shown in Table~\ref{tab_bench}.  One way to help release this issue is to go for a top-down approach instead (Section~\ref{sec_topdown}).%

\vspace*{4mm}\section{Top-down design approach -- Trading off task accuracy and efficiency} \label{sec_topdown}

The top-down neuromorphic design approach attempts at answering the key difficulty of bottom-up designs in tackling real-world problems efficiently, beyond neuroscience-oriented applications (Fig.~\ref{fig_strategy}). Taking inspiration from the field of dedicated machine-learning accelerators, top-down design (i)~starts from the applicative problem and the related algorithms, (ii)~investigates how to release key constraints in order to make these algorithms hardware- and biophysics-aware, and (iii)~proceeds with the hardware integration. This leads to a tradeoff between \textit{efficiency} and \textit{accuracy} on the selected use case. The resulting designs can thus be distinguished from their bottom-up counterparts studied in Section~\ref{sec_bottomup} in that they can hardly be applied to another purpose than the one they were designed and optimized for (e.g.,~speech instead of image recognition), although recent developments may help release this restriction (see Section~\ref{sec_discussion}).

Interestingly, in line with the challenge of embedded synaptic plasticity highlighted by bottom-up approaches, edge computing research currently sees the integration of on-chip learning capabilities within power budgets of sub- to tens of $\mu$W as one of the next grand challenges~\cite{Murmann20}. Therefore, we will now focus on algorithmic aspects linked to an error-based training of spiking neural networks, in direct contrast with bottom-up synaptic plasticity aspects discussed in Section~\ref{sec_bottomup}. Following the steps of the top-down approach (Fig.~\ref{fig_strategy}), we first cover SNN training algorithms in Section~\ref{ssec_algos} and then move to their silicon implementations in Section~\ref{ssec_siac}.

\subsection{Algorithms}\label{ssec_algos}

The backpropagation of error (BP) algorithm~\cite{Rumelhart86,Schmidhuber15} is usually chosen as a starting point for SNN training, however it needs to be adapted due to the non-differentiable nature of the spiking activation function. In this respect, several techniques were proposed, such as linearizing the membrane potential at the spike time~\cite{Bohte02}, temporally convolving spike trains and computing with their differentiable smoothened version~\cite{Mohemmed12}, treating spikes and discrete synapses as continuous probabilities from which network instances can be sampled~\cite{Esser15}, treating the influence of discontinuities at spike times as noise on the membrane potential~\cite{Lee16}, using a spiking threshold with a soft transition~\cite{Huh18}, or differentiating the continuous spiking probability density functions instead of discontinuous membrane voltage traces~\cite{Shrestha18}. Another popular and robust approach consists in using a \textit{surrogate gradient} in place of the spiking activation function derivative during the backward pass~\cite{Zenke18,Neftci19,Zenke21}, similarly to the use straight-through estimators for non-differentiable activation functions in ANNs~\cite{Bengio13,Courbariaux16,Hubara18}, which is increasingly being supported through open-source toolboxes such as Norse~\cite{Pehle21}, SpikingJelly~\cite{Fang20} and snnTorch~\cite{Eshraghian21}.

\begin{figure}[!t]
\centering
\noindent\hspace*{-5mm}\includegraphics[width=0.58\columnwidth]{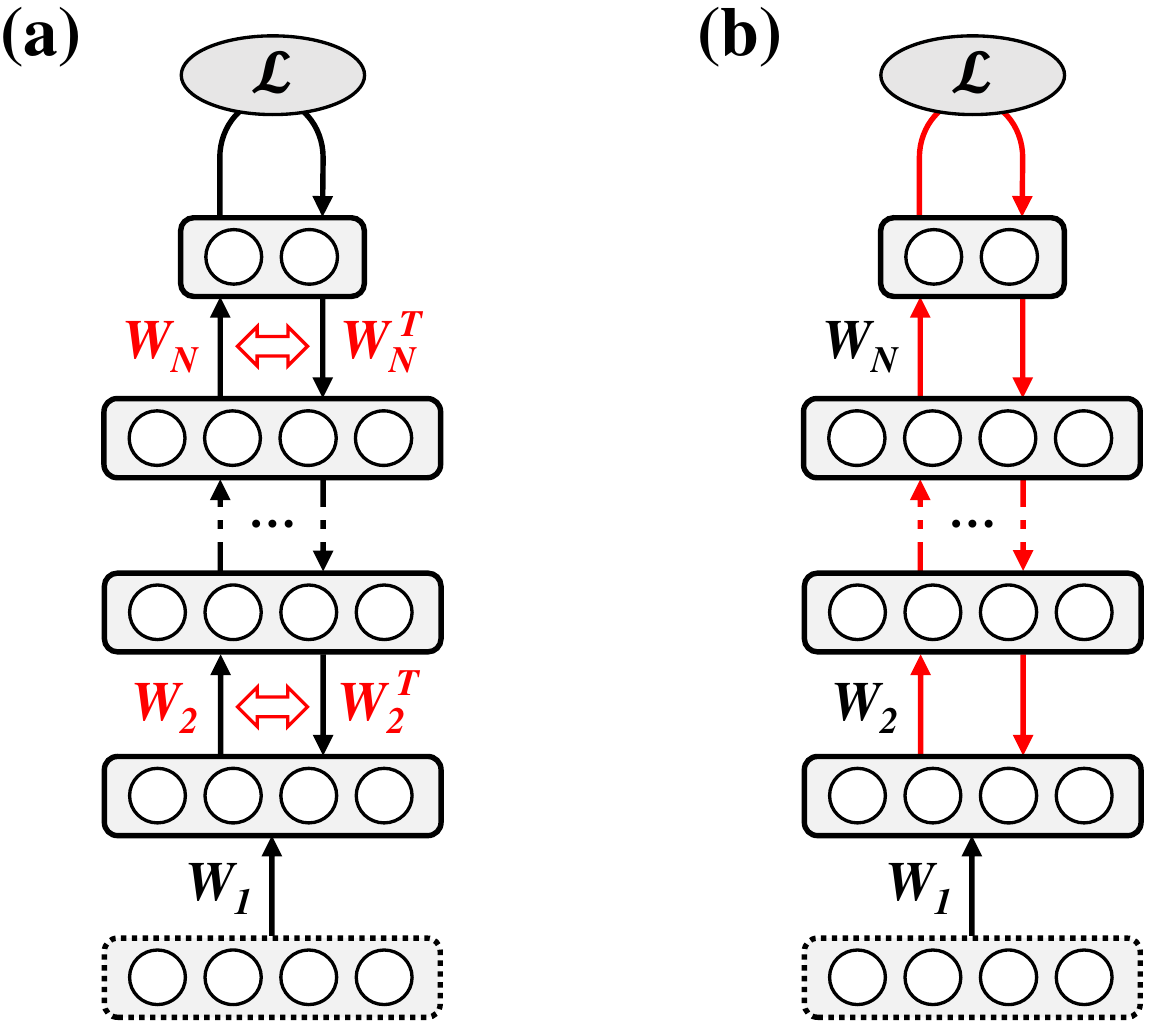}
\caption{Illustration, on an $N$-layer network, of the two key challenges of the backpropagation of error algorithm, which impact both biological plausibility and hardware efficiency: \textbf{(a)}~\textit{the weight transport problem}, which requires accessing the weight matrices in the forward pass and their transpose in the backward pass, and \textbf{(b)}~\textit{update locking}, which requires carrying out the forward and backward passes entirely before the weights of the first hidden layer ($W_1$) can be updated.}%
\vspace*{-1mm}
\label{fig_BPissues}
\end{figure}

However, while these techniques allow for the application of BP to SNNs, it is also necessary to reduce the computational complexity and memory requirements of BP toward an on-chip implementation. The first key issue of BP is the \textit{weight transport problem}, also known as \textit{weight symmetry}~\cite{Grossberg87,Liao16} and illustrated in Fig.~\ref{fig_BPissues}(a): the same weight values need to be accessed during the forward and the backward passes, implying the use of complex memory access patterns and architectures. The second key issue of BP is \textit{update locking}~\cite{Jaderberg17,Czarnecki17}, shown in Fig.~\ref{fig_BPissues}(b), which entails severe memory and latency overheads as it requires (i) buffering the activation values of all layers, and (ii) carrying out the full forward and backward passes before the weights of the first hidden layer can be updated (see $W_1$ in Fig.~\ref{fig_BPissues}(b)). Interestingly, these issues also preclude BP from being biologically plausible~\cite{Bengio15}, and both of them arise from a non-locality of error signals and weights during the forward and backward passes~\cite{Neftci18}. On the one hand, locality of the error signals can be addressed with layerwise loss functions allowing for an independent training of the layers with local error information~\cite{Mostafa18,Nokland19,Kaiser20}. A similar strategy is pursued in \textit{synthetic gradient} approaches~\cite{Jaderberg17,Czarnecki17}, which rely on local gradient predictors. Yet another approach consists in defining target values based on layerwise auto-encoders~\cite{Lee15,Meulemans20}. On the other hand, approaches aiming at weight locality are found in the recent development of \textit{feedback-alignment}-based algorithms~\cite{Lillicrap16,Baldi18,Nokland16,FrenkelLefebvre21}. They rely on fixed random connectivity matrices in the error pathway, either as a direct replacement of the backward weights (feedback alignment, FA~\cite{Lillicrap16,Baldi18}), for a projection of the network output error on a layerwise basis (direct feedback alignment, DFA~\cite{Nokland16}), or for a projection of the one-hot-encoded classification labels (direct random target projection, DRTP~\cite{FrenkelLefebvre21}). Interestingly, the DRTP algorithm releases not only the weight transport problem, but also update locking by ensuring locality in both weights and error signals. However, feedback-alignment-based algorithms currently do not offer a satisfactory performance for the training of convolutional neural networks (CNNs) as the convolutional kernel weights have insufficient parameter redundancy, which is known as the \textit{bottleneck effect}~\cite{Lillicrap16,Launay19,FrenkelLefebvre21}.

The above-mentioned algorithms can be straightforwardly applied to SNNs with rate-based coding. For example, DFA has been formulated as a three-factor rule for SNNs in~\cite{Neftci17}, and DECOLLE was shown to be suitable for memristive neuromorphic hardware in~\cite{Payvand20}. However, rate-based coding implies two key issues. First, updates cannot be carried out as long as activity has not reached a steady-state regime, leading to a latency penalty~\cite{Neftci17}. Second, due to its non-sparse nature where every spike only contains a marginal amount of information, rate coding is unlikely to lead to any power advantage compared to conventional non-spiking approaches, as shown in~\cite{Davidson21}. This issue also applies to ANN-to-SNN mapping approaches that rely on the equivalence between the ReLU activation function and the spike rate of an I\&F neuron~\cite{Diehl15,Diehl16,Rueckauer17}. Therefore, taking time into consideration is necessary, otherwise the key opportunities in sparsity and low power consumption of SNNs cannot be exploited. To solve this issue, several gradient-based algorithms exploiting a TTFS encoding were proposed~\cite{Mostafa17,Kheradpisheh20,GoltzKriener19}. The algorithm from~\cite{GoltzKriener19} was demonstrated with the BrainScaleS-2 system for variability-aware training, although based on a setup where an off-chip optimizer retrieves state and activation data online from a BrainScaleS-2 chip. This chip-in-the-loop setup was selected as the full update rules have a complexity level that is incompatible with an on-chip implementation. However, a simplified version was also shown in~\cite{GoltzKriener19} to exhibit a low complexity while maintaining the learning ability on simple tasks.

\begin{figure}[!t]
\centering
\noindent\includegraphics[width=1.00\columnwidth]{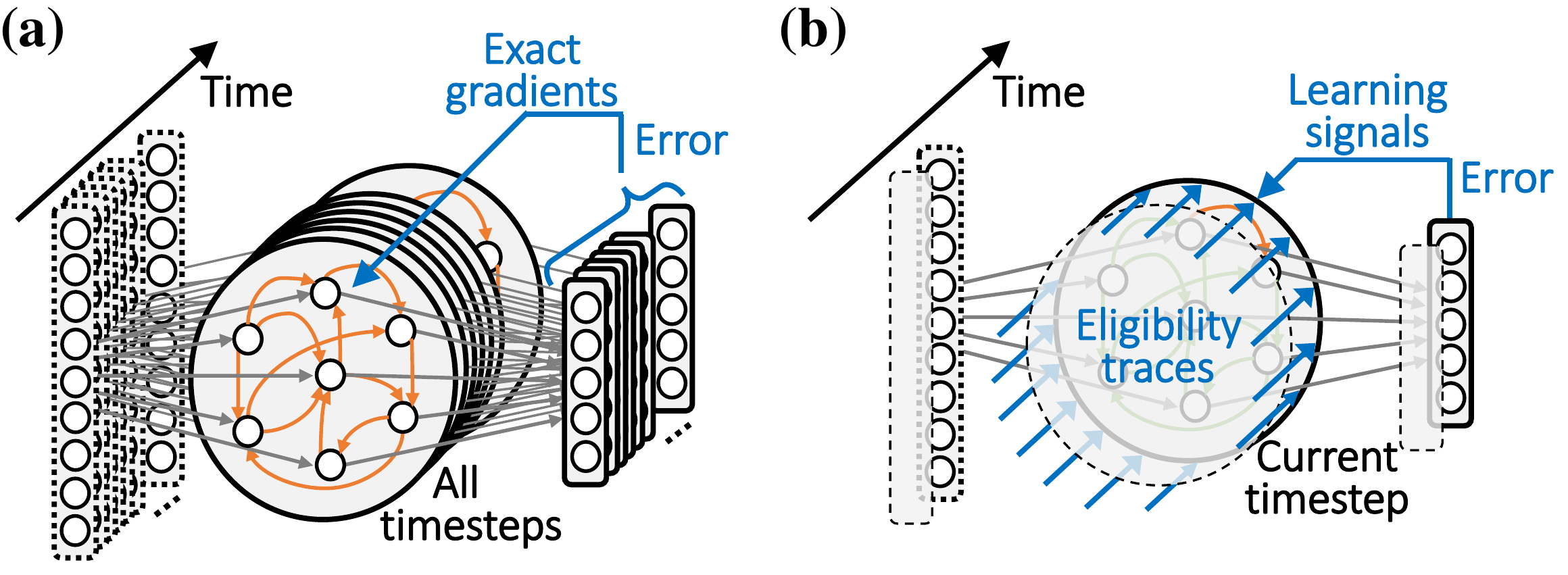}
\caption{Overview of the training strategies that can be used to learn from temporal data, illustrated for a single-layer recurrent neural network. \textbf{(a)}~\textit{Backward-mode learning}, as per the standard backpropagation through time (BPTT) algorithm~\cite{Werbos90}. \textbf{(b)}~\textit{Forward-mode learning}, illustrated based on the terminology introduced with the e-prop and OSTL algorithms~\cite{Bellec20,Bohnstingl20}: learning signals are an error-dependent term available locally in time, while eligibility traces are non-error-dependent terms that are computed online and capture the importance of each synapse on the network output. Adapted from~\cite{Frenkel22}.}%
\label{fig_backforward}
\end{figure}

In order to perform gradient-based training in both space and time with recurrent neural networks (RNNs), another approach consists in starting from the backpropagation through time (BPTT) algorithm~\cite{Werbos90}. However, BPTT requires \textit{unrolling the network in time} in order to backpropagate error gradients through the network dynamics (see Fig.~\ref{fig_backforward}(a) for an illustration), which leads to intractable memory requirements for low-power hardware. Approximations of BPTT were thus investigated, among which the e-prop~\cite{Bellec20} and the online spatio-temporal learning (OSTL)~\cite{Bohnstingl20} algorithms. The former relies on the simplification that only the direct influence of spikes on the output error is taken into account, not their influence on future errors through the network dynamics. The latter elegantly separates the spatial and temporal components of the gradient, and approximates to zero a residual term resulting from cross-layer spatio-temporal dependencies. As shown in Fig.~\ref{fig_backforward}(b), the weight updates of both algorithms can be formulated as the product between a \textit{learning signal}, an error-dependent term available locally in time, and an \textit{eligibility trace}, which is computed online and is a biologically plausible primitive (see Section~\ref{sssec_synapses}). Importantly, both \mbox{e-prop} and OSTL can be applied \textit{online} as new data is provided (i.e.~no unrolling of the network in time is required): they belong to the class of \textit{forward-mode} learning algorithms, in contrast with BPTT that is a \textit{backward-mode} learning algorithm. They can be seen as simplifications of the real-time recurrent learning (RTRL) algorithm, which was proposed in 1989 as the first forward-mode alternative to BPTT~\cite{Williams89}, but whose prohibitive memory and time complexities have precluded its adoption in its original form~\cite{ZenkeNeftci21}. Both e-prop and OSTL have been applied successfully to spiking RNNs~\cite{Bellec20,Bohnstingl20}, while another forward-mode learning algorithm known as forward propagation through time (FPTT)~\cite{Kag21} was also successfully applied to SNNs in~\cite{Yin21}.

\begin{table*}
\hspace*{1mm}\begin{minipage}{.98\textwidth}
\vspace*{-3.7mm}
\caption{Comparison of top-down neuromorphic chips. The three designs on the right combine bottom-up and top-down approaches.}
\label{table_SoA_topDown}
\renewcommand{\arraystretch}{1.05}
\centering
\resizebox{\textwidth}{!}{\begin{tabular}{lccccccccccc}%
\toprule%
Author & ~~~~ & Knag~\cite{Knag15} & Kim~\cite{Kim15} & Buhler~\cite{Buhler17} &  Park~\cite{Park19} & Frenkel~\cite{Frenkel20} & Frenkel~\cite{Frenkel22} & ~~~~ & Chen~\cite{Chen18} & Pei~\cite{Pei19} & Neckar~\cite{Neckar19} \\
Publication & & JSSC, 2015 & VLSI-C, 2015 & VLSI-C, 2017 & JSSC, 2019 & ISCAS, 2020 & ISSCC, 2022 & & JSSC, 2019 & Nature, 2019 & PIEEE, 2019 \\
Chip name & & -- & -- & -- & -- & SPOON & ReckOn & & -- & Tianjic & Braindrop \\\midrule

Implementation & & Digital & Digital & Mixed-signal & Digital & Digital & Digital & & Digital & Digital & Mixed-signal \\

Technology & & 65\,nm & 65\,nm & 40\,nm & 65\,nm & 28\,nm FDSOI & 28\,nm FDSOI & & 10\,nm FinFET & 28\,nm HPL & 28\,nm FDSOI \\%

Pad-free area & & 3.1\,mm$^2$ & 1.8\,mm$^2$ & 1.3\,mm$^2$ & 10.1\,mm$^2$ & 0.32\,mm$^2$ & 0.45\,mm$^2$ & & 1.72\,mm$^2$ & 14.4\,mm$^2$ & 0.65\,mm$^2$ \\

Architecture & & Spiking LCA & Spiking LCA & Spiking LCA & BNN & eCNN & Spiking RNN & & SNN/BNN & SNN/ANN & SNN \\

\multirow{2}{*}{\shortstack{Topology \\ \# syn}} & & \multirow{2}{*}{\shortstack{4$\times$64\\128k (8,13-bit)}} & \multirow{2}{*}{\shortstack{4$\times$64\\83k (4,5,14-bit)}} & \multirow{2}{*}{\shortstack{8$\times$64\\N/A}} & \multirow{2}{*}{\shortstack{(784)--200--200--10\\194k (14-bit)}} & \multirow{2}{*}{\shortstack{C5$\times$5@10--128--10\\64k (8-bit)}}  & \multirow{2}{*}{\shortstack{(256)--R256--16\\132k (8-bit)}} & & \multirow{2}{*}{\shortstack{64$\times$64\\1M (7-bit)}} & \multirow{2}{*}{\shortstack{156$\times$256 \\10M (8-bit)}} & \multirow{2}{*}{\shortstack{4k \\64k (8-bit)$^\diamond$}} \\ \\

\multirow{3}{*}{\shortstack{Embedded\\online\\learning}} & & \multirow{3}{*}{\shortstack{SAILnet\\(unsupervised)}} & \multirow{3}{*}{\shortstack{BP\\(last layer only)}} & \multirow{3}{*}{\shortstack{\textit{Yes}\\\textit{(unspecified)}}} & \multirow{3}{*}{Mod. DFA} & \multirow{3}{*}{DRTP} & \multirow{3}{*}{\shortstack{Mod. stoch. e-prop}} & & \multirow{3}{*}{STDP} & \multirow{3}{*}{No} & \multirow{3}{*}{No}\\ \\ \\

\multirow{3}{*}{\shortstack{Demonstrated\\application}} & & \multirow{3}{*}{\shortstack{Image sparse\\coding}} & \multirow{3}{*}{\shortstack{Image sparse\\coding \& recog.}} & \multirow{3}{*}{\shortstack{Image sparse\\coding \& recog.}} & \multirow{3}{*}{Image recog.} & \multirow{3}{*}{Image recog.} & \multirow{3}{*}{\shortstack{Gesture recog.,\\Keyword spotting,\\Navigation}} & & \multirow{3}{*}{\shortstack{Image sparse\\coding \& recog.}} & \multirow{3}{*}{\shortstack{Real-time image,\\sound recognition\\\& control}} & \multirow{3}{*}{\shortstack{NEF-based\\networks}}\\ \\ \\

\multirow{3}{*}{Benchmark(s)$^\ddag$} & & \multirow{3}{*}{Denoising} & \multirow{3}{*}{MNIST (\textbf{84\%--90\%})} & \multirow{3}{*}{MNIST (\textbf{88\%})} & \multirow{3}{*}{MNIST (\textbf{97.8\%})} & \multirow{3}{*}{\shortstack{MNIST (\textbf{95.3\%},97.5\%),\\N-MNIST (\textbf{93.0\%},93.8\%)}} & \multirow{3}{*}{\shortstack{IBM DVS Gest. (\textbf{87.3\%}),\\SH Digits (\textbf{90.7\%}),\\Delayed cue acc. (\textbf{96.4\%})}} & & \multirow{3}{*}{\shortstack{Denoising,\\MNIST (98.6\%)}} & \multirow{3}{*}{\shortstack{Autonomous\\ bike driving$^*$}} & \multirow{3}{*}{\shortstack{Function fitting,\\integrator}} \\ \\  \\ 

\multirow{2}{*}{Energy metric} & & \multirow{2}{*}{48\,pJ/pix} & \multirow{2}{*}{5.7\,pJ/pix} & \multirow{2}{*}{48.9\,pJ/pix} & \multirow{2}{*}{302\,pJ/pix} & \multirow{2}{*}{\shortstack{1.7\,nJ per\\pixel event$^\dag$}} & \multirow{2}{*}{5.3\,pJ/SOP} & & \multirow{2}{*}{3.8\,pJ/SOP} & \multirow{2}{*}{\shortstack{0.78\,pJ/OP,\\1.54\,pJ/SOP}} & \multirow{2}{*}{0.38\,pJ/SOP}\\ \\

\bottomrule
\end{tabular}}
\vspace*{1.5mm}
\begin{adjustwidth}{0.3cm}{}
\begin{spacing}{0.5}
{\flushleft\fontsize{5.8pt}{6pt}\selectfont$^\ddag$ Accuracy results in bold font are obtained with on-chip online learning.}
\end{spacing}
\end{adjustwidth}
\vspace*{-2mm}
\begin{adjustwidth}{0.3cm}{}
\begin{spacing}{0.5}
{\flushleft\fontsize{5.8pt}{6pt}\selectfont$^\dag$ Pre-silicon results.}
\end{spacing}
\end{adjustwidth}
\vspace*{-2mm}
\begin{adjustwidth}{0.3cm}{}
\begin{spacing}{0.5}
{\flushleft\fontsize{5.8pt}{6pt}\selectfont$^*$ Pei~\textit{et~al.}~also use N-MNIST and MNIST to quantify the efficiency and throughput improvement over a GPU and the improvement brought by hybrid SNN-ANN processing over SNN-only processing, respectively. However, the reported results are used only for relative comparisons, the provided data is not sufficient to be included in this table and in Section~\ref{sssec_topdown_tradeoff}.}
\end{spacing}
\end{adjustwidth}
\vspace*{-2mm}
\begin{adjustwidth}{0.3cm}{}
\begin{spacing}{0.5}
{\flushleft\fontsize{5.8pt}{6pt}\selectfont$^\diamond$ Refers to the weight resolution. The effective number of bits per synapse is typically lower and depends on the implemented network topology.}
\end{spacing}
\end{adjustwidth}
\vspace*{-2.5mm}
\end{minipage}
\end{table*}

Just as the latter BPTT-derived rules can be mapped onto bio-plausible synaptic eligibility traces, there is a growing interest into the development of algorithms that can be mapped onto primitives related to dendritic processing. In~\cite{Guerguiev17}, Guerguiev~\textit{et~al.} show how segregated apical and basal dendritic compartments can be used to integrate feedback and feedforward signals, respectively (see Fig.~\ref{fig_introNeurons}). However, it does so in two distinct \textit{forward} and \textit{target} phases, which is not biologically plausible and entails update locking. This constraint is released in the cortical model proposed by Sacramento~\textit{et~al.}: apical dendrites encode prediction errors resulting from top-down network-level feedback and modulate, through the soma, the plasticity of synapses located on basal dendrites, which receive feedforward sensory input~\cite{Sacramento18}. This model is based on the concept of predictive coding outlined in Section~\ref{sssec_dendrites} and is closely related to the work~\cite{Whittington17}, where prediction errors are represented in specific subpopulations of neurons instead of dendrites. Importantly, the work of Payeur~\textit{et~al.} demonstrates how to combine numerous bio-inspired elements mentioned in Section~\ref{sec_bottomup}, such as bursts of spikes, voltage traces, dendritic compartments, neuromodulation and STP~\cite{Payeur21}. For the first time, scaling to machine learning datasets as complex as ImageNet~\cite{Deng09} is demonstrated. Although this scaling is still at a proof-of-concept level with an inefficient resource usage, this is a key first step toward large-scale bio-plausible learning.

Finally, for energy-based models (of which Hopfield networks may be the prime example~\cite{Hopfield82}), the equilibrium propagation algorithm offers an alternative to BPTT for an implementation of gradient-based training~\cite{Scellier17}. While BPTT requires carrying out distinct computations in the forward and backward passes of the algorithm, equilibrium propagation estimates gradients by running the energy-based model in two phases: a \textit{free phase} until the network reaches equilibrium, and a \textit{nudging phase} during which the output neurons are nudged toward the desired solution, leading to a new equilibrium. Updates can then be carried out based on the results of these two phases. As this would lead to hardware constraints similar to those of update locking, another version of the equilibrium propagation algorithm has been proposed in which weights can be updated in a continuous manner during the nudging phase~\cite{Ernoult20}. This continuous version recently led to a first spike-based implementation of equilibrium propagation in~\cite{Martin21}. However, the use of rate coding currently implies latency and power penalties similar to those of the previously-mentioned DFA-based and DECOLLE-based spiking algorithms of~\cite{Neftci17} and~\cite{Payvand20}, respectively.

\subsection{Silicon implementation}\label{ssec_siac}

While most of the algorithms outlined in Section~\ref{ssec_algos} result from recent developments, some of them already made it to silicon. We first survey top-down designs qualitatively to illustrate their applicative landscape, including developments merging bottom-up and top-down insight (Section~\ref{sssec_topdown_overview}). We then quantitatively assess the key accuracy/efficiency tradeoff that top-down designs optimize for their selected use cases (Section~\ref{sssec_topdown_tradeoff}).

\vspace*{3mm}\subsubsection{Overview of neuromorphic accelerators}\label{sssec_topdown_overview}~\\\vspace*{-3.5mm}

As the scopes, implementations and applications of top-down designs vary widely, comparing them directly is difficult, except when standard benchmarks are used. In order to extract the main trends, a summary of top-down neuromorphic designs is provided in Table~\ref{table_SoA_topDown}. 

The three chips from Knag~\textit{et~al.}~\cite{Knag15}, Kim~\textit{et~al.}~\cite{Kim15} and Buhler~\textit{et~al.}~\cite{Buhler17} follow a similar approach: they enforce a sparse feature representation of input images by introducing competition between groups of neurons (i.e.~locally competitive algorithm, LCA). The LCA is implemented as a systolic ring of SNN cores, each of which is fully connected to input pixels with feedforward excitatory connections, while lateral connections between neurons are inhibitory to favor sparsity in image representation. The 65-nm digital chip from Knag~\textit{et~al.}~furthermore implements SAILnet, a bio-inspired unsupervised algorithm with local spike-based plasticity for adaptation of the neuron receptive fields~\cite{Zylberberg11}. Its main purpose is thus image feature extraction applied to denoising, however it lacks an inference module for image recognition and classification. This point is addressed by the chips from Kim~\textit{et~al.}~and Buhler~\textit{et~al.} The former is a 65-nm digital design whose last layer can be trained with stochastic gradient descent (SGD) to perform classification. The latter is a 40-nm mixed-signal design embedding analog LIF neurons, it is also claimed to embed online learning, but without specifying the associated algorithm. Both chips are benchmarked on MNIST~\cite{LeCun98}, although with limited accuracies ranging from 84\% to 90\%.

Another approach is proposed by Park~\textit{et~al.}~\cite{Park19}, whose claim is to leverage the advantages of both ANNs (i.e.~single-timestep frame-based processing) and SNNs (i.e.~sparse binary activations). The proposed architecture is thus equivalent to a binary neural network (BNN). It embeds the bio-inspired version of the DFA algorithm proposed by Guerguiev~\textit{et~al.}~in~\cite{Guerguiev17}. Although DFA suffers from update locking, which implies a pipelined weight update scheme, Park~\textit{et~al.}~demonstrate a low-power design achieving an accuracy of 97.8\% on MNIST with on-chip online learning.

Therefore, top-down neuromorphic designs mostly split among two categories: SNNs with event-driven processing at the expense of accuracy~\cite{Knag15,Kim15,Buhler17} or BNNs with high accuracy at the expense of conventional frame-based processing~\cite{Park19}. The SPOON chip proposed in~\cite{Frenkel20} aims at bridging the two approaches. It is a 28-nm event-driven CNN (eCNN) combining both event-driven and frame-based processing: through the use of a TTFS code, the former leverages sparsity from event-based neuromorphic sensors~\cite{Lichsteiner08,Posch11,Brandli14,Vanarse16}, while the latter ensures efficiency, accuracy and low latency during training and inference. It also embeds the low-overhead DRTP algorithm in its fully-connected layers. SPOON is benchmarked on MNIST and on the spike-based neuromorphic MNIST (N-MNIST) dataset~\cite{Orchard15b}, which was generated by presenting MNIST images to a neuromorphic vision sensor~\cite{Posch11} mounted on a pan-tilt unit moved in three saccades. SPOON reaches accuracies of 95.3\% (on-chip training) and 97.5\% (off-chip training) on MNIST, and of 93.0\% (on-chip training) and \mbox{93.8\% (off-chip training) on N-MNIST}. 

To go beyond static images or temporally-coded static data, the ReckOn chip proposed in~\cite{Frenkel22} endows a spiking RNN architecture with the ability to learn over second-long timescales. This is achieved through a modified version the e-prop algorithm, where eligibility traces were adapted to scale with the number of neurons instead of synapses for a low-cost solution that consumes less than 50-$\mu$W for real-time learning with a silicon core area of 0.45\,mm$^2$. The code-agnostic learning property of e-prop is combined with the sensor-agnostic property of spike-based encodings for end-to-end on-chip task-agnostic learning, which was demonstrated on vision (87.3-\% accuracy gesture recognition), audition (90.7-\% accuracy keyword spotting), and navigation (96.4-\% accuracy behavioral-timescale cue accumulation) tasks. We further discuss temporal-data benchmarks in Section~\ref{ssec_challenges}.

Finally, three recently-published chips highlight that embedding bottom-up insight into a top-down approach can be beneficial to neuromorphic computing (Table~\ref{table_SoA_topDown}): the chip from Chen~\textit{et~al.}~\cite{Chen18}, Tianjic~\cite{Pei19} and Braindrop~\cite{Neckar19}. The first one is another attempt to bridge the gap between the BNN and SNN trends with a low-power STDP-based SNN in 10-nm FinFET that can also be programmed as a BNN. However, these two modes are still segmented at the application level: SNN operation with STDP is chosen for image denoising and BNN operation with offline-trained weights is chosen for image recognition. Indeed, Chen~\textit{et~al.} show that an offline-trained BNN achieves 98.6\% on MNIST, while a single-layer SNN with STDP training only achieves 89\% on a pre-processed Gabor-filtered version of MNIST. Event-driven computation can thus not be leveraged in this device if high accuracy is required. The second one is Tianjic, a 28-nm digital design allowing for hybrid ANN-SNN setups and embedding as high as 40k neurons and 10M synapses per chip. This scale allows multi-chip Tianjic setups to be benchmarked on an autonomous bike driving task, demonstrating how both the ANN and SNN paradigms can be combined for real-time image recognition, sound recognition, and vehicle control. The third one is Braindrop, a 28-nm mixed-signal design that relies, together with its software frontend, on an efficient set of mismatch- and temperature-invariant abstractions to provide one-to-one correspondence with the neural engineering framework (NEF)~\cite{Eliasmith04} (see also Section~\ref{ssec_challenges}). It follows an encode-transform-decode architecture directly inspired by the previous-generation bottom-up Neurogrid design~\cite{Benjamin14}, and was benchmarked on nonlinear 1D and 2D function fitting tasks and on integrator modeling. These three chips demonstrate a high energy efficiency with 3.8\,pJ/SOP for the chip of Chen~\textit{et~al.}, 0.78\,pJ/OP (ANN setup) or 1.54\,pJ/SOP (SNN setup) for Tianjic and 0.38\,pJ/SOP for Braindrop. However, Braindrop and Tianjic do not embed online learning and require an offline setup for network training and programming, while the STDP rule in the chip from Chen~\textit{et~al.} has a limited training ability beyond denoising tasks (Table~\ref{tab_bench}).

\begin{figure}%
\centering%
\vspace*{-1mm}
\includegraphics[width=0.975\columnwidth]{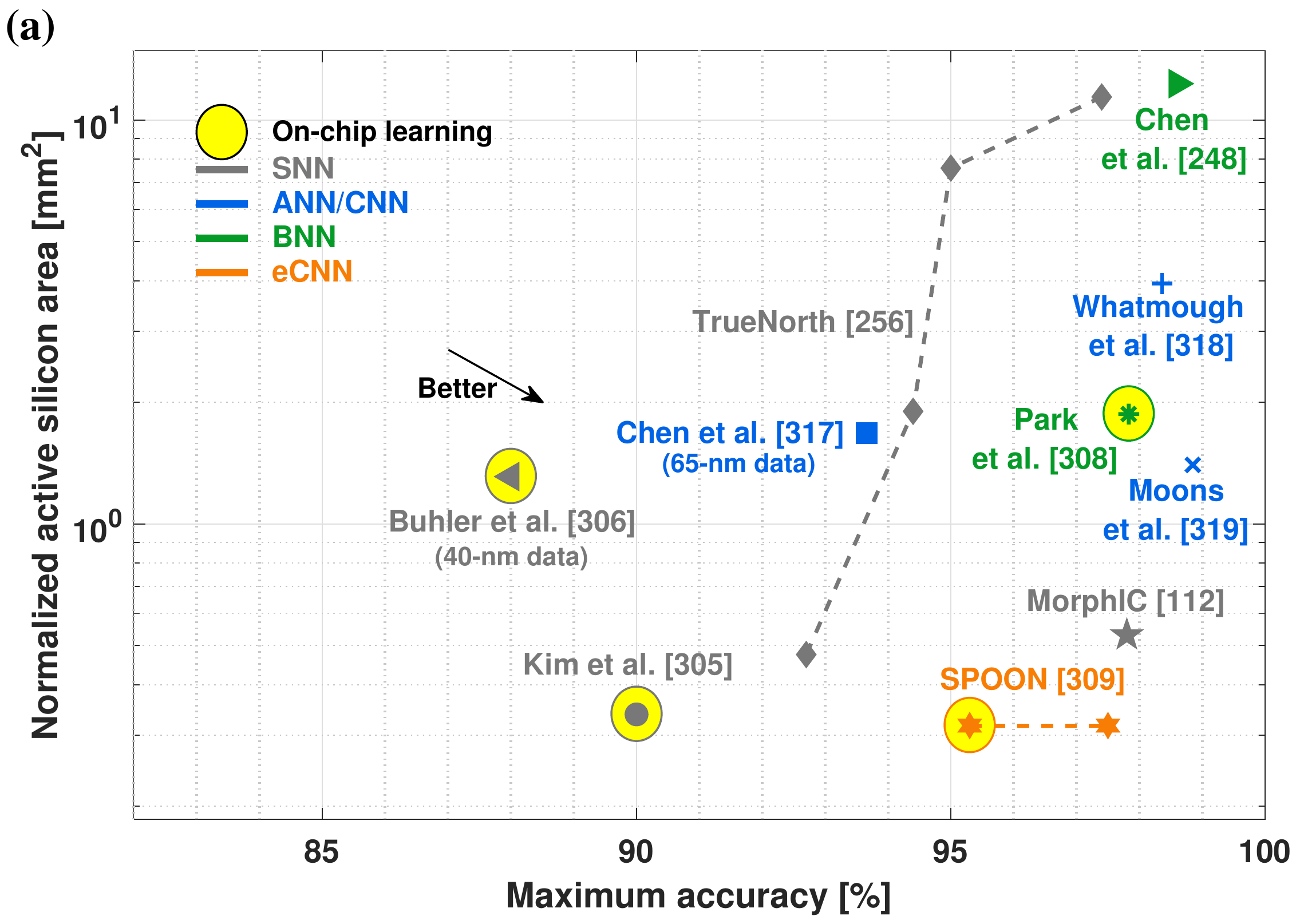}\\%
\includegraphics[width=0.975\columnwidth]{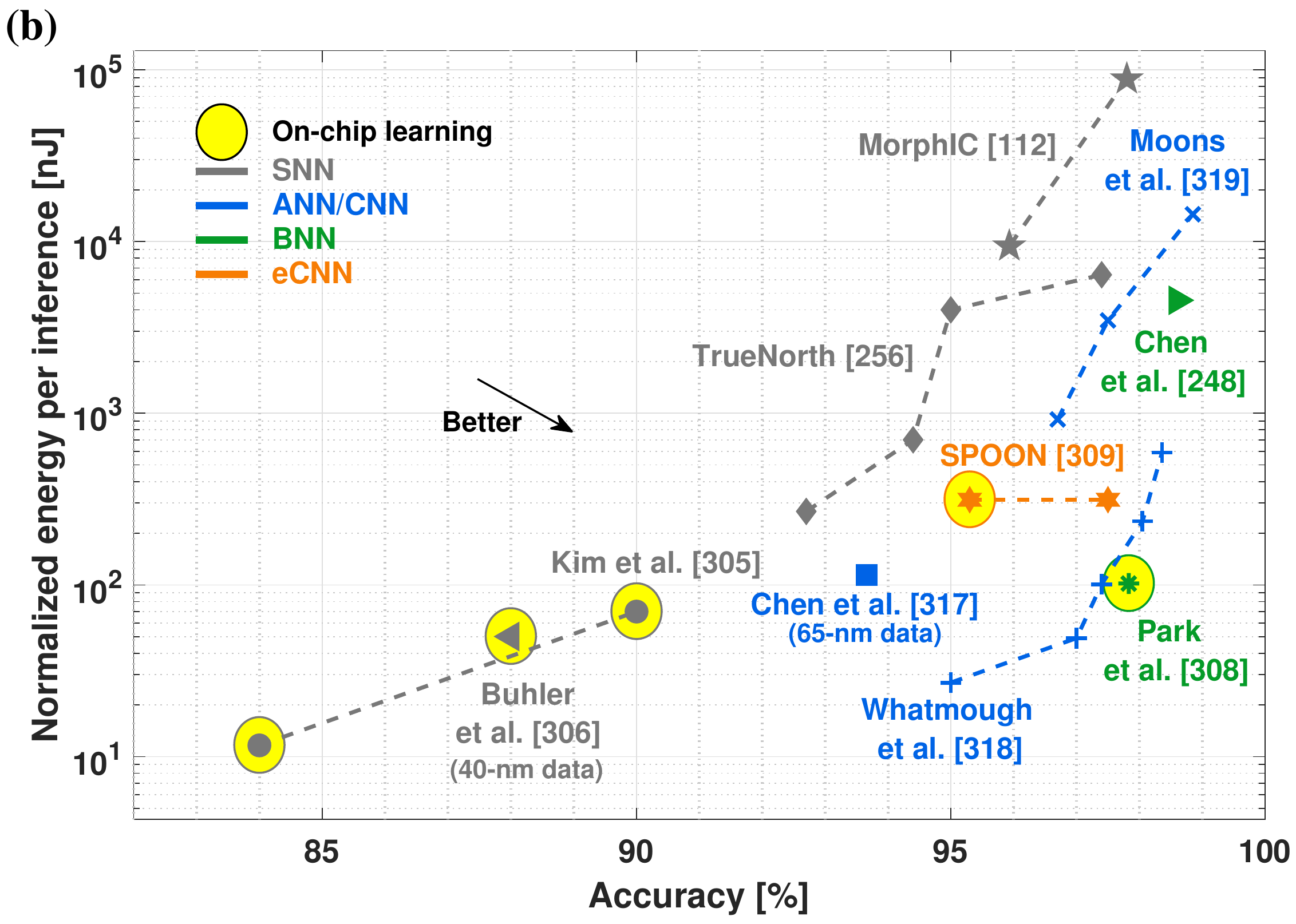}%
\vspace*{-1.5mm}
\caption{Analysis of tradeoffs between accuracy, area and energy per classification on the MNIST dataset for SNNs, BNNs, ANNs and CNNs, where results obtained on pre-processed or simplified versions of MNIST have been excluded. Although MorphIC and the chip from Chen~\textit{et~al.} embed online learning, the MNIST experiments of these two chips were obtained with offline-learned weights. Results on the non-pre-processed MNIST dataset are reported for the chip from Chen~\textit{et~al.} in its BNN configuration. All chips are digital and allow for technology normalization, except the 40-nm design from Buhler~\textit{et~al.} and the 65-nm design from Chen~\textit{et~al.}, which are mixed-signal~\cite{Buhler17,Chen19}. Pre-silicon results are reported for SPOON. (\textbf{a}) Area-accuracy tradeoff. Silicon area (excluding pads) has been  normalized to a 28-nm technology node using the node factor (e.g., a (28/65)$^2$-fold reduction for normalizing 65\,nm to 28\,nm), except for the 10-nm FinFET node from Chen~\textit{et~al.}~\cite{Chen18} where data from~\cite{Mistry17} was used for normalization. The TrueNorth area varies as Esser~\textit{et~al.}~used different numbers of cores for their experiments (5, 20, 80 and 120 cores, in the order of increasing accuracy)~\cite{Esser15}. A 1920-core configuration is also reported in~\cite{Esser15}, leading to a 99.42-\% accuracy on MNIST with TrueNorth, a record for SNNs. However, as this configuration would lead to a normalized area of 980\,mm$^2$, we only included TrueNorth configurations whose scale are comparable with the other chips. (\textbf{b}) Energy-accuracy tradeoff. Energy has been normalized to a 28-nm technology node using the node factor (e.g., a (28/65)-fold reduction for normalizing 65\,nm to 28\,nm). Adapted from~\cite{Frenkel20}.}
\vspace*{-3mm}
\label{fig_ppa}
\end{figure}

\vspace*{3.5mm}\subsubsection{Accuracy / efficiency comparative analysis}\label{sssec_topdown_tradeoff}~\\\vspace*{-1mm}

While bottom-up SNN designs favor a comparison based on low-level criteria such as neuron behaviors, synaptic plasticity and weight resolution, neuron and synapse densities, energy per SOP, or fan-in and fan-out (Section~\ref{sssec_bottomup_tradeoff}), top-down neuromorphic approaches require a comparison based on benchmark performance as they start from the applicative problem. Currently, MNIST is the only dataset for which data is available for many bottom-up and top-down neuromorphic designs, as well as for conventional machine-learning accelerators. Therefore, MNIST allows for accuracy/efficiency comparisons across all neural network types, including SNNs, BNNs, ANNs and CNNs (see further discussion in Section~\ref{ssec_challenges}). 

The tradeoff analysis of energy, area and accuracy on the MNIST dataset is shown in Fig.~\ref{fig_ppa}, which has been normalized to a 28-nm technology node to allow for fair comparisons, except for the two mixed-signal designs of~\cite{Buhler17,Chen19}. SNNs appear to lag behind conventional ANN and CNN accelerators~\cite{Whatmough17,Moons18}, the BNN from Park~\textit{et~al.}~\cite{Park19}, the chip from Chen~\textit{et~al.} in its BNN configuration~\cite{Chen18}, and the SPOON eCNN~\cite{Frenkel20}. Among SNNs, MorphIC achieves a high area efficiency without incurring a power penalty.~Interestingly, the hybrid approach pursued in SPOON leads to the only design achieving the efficiency of conventional machine-learning accelerators while enabling online learning with event-based sensors, thanks to a tight combination of event-driven and frame-based processing supported by DRTP on-chip training. Similar trends were also recently outlined in Tianjic by Pei~\textit{et~al.}, where a hybrid ANN-SNN network was demonstrated to outperform the equivalent SNN-only network~\cite{Pei19}. These findings form an interesting trend worth investigating for the deployment of top-down neuromorphic designs in real-world applications.\\

\section{Discussion and outlook} \label{sec_discussion}
\vspace*{3mm}

From this comprehensive overview of the bottom-up and top-down neuromorphic design approaches, it is possible to identify important synergies. In the following, we discuss them toward the goal of neuromorphic intelligence (Section~\ref{ssec_trends}), elaborate on the missing elements and open challenges (Section~\ref{ssec_challenges}), and finally outline some of the most promising use cases (Section~\ref{ssec_appli}).

\subsection{Merging the bottom-up and top-down design approaches}\label{ssec_trends}

The \textit{science}-driven bottom-up design approach, which aims at replicating and understanding \textit{natural intelligence}, is driven mainly by neuroscience observations, under the constraint of optimizing the silicon implementation efficiency of neuron versatility, synaptic plasticity, and communication infrastructure scalability. Through Section~\ref{sec_bottomup}, we highlighted how these tradeoffs can be optimized \textit{in silico}, but also showed that bottom-up designs can struggle to achieve the efficiency of dedicated machine-learning accelerators at iso-accuracy. Identifying suitable applications that can exploit the design choices driven by neuroscience considerations and lead to a competitive advantage over conventional approaches is still an open challenge. 

The \textit{engineering}-driven top-down design approach, which aims at designing \textit{artificial intelligence} devices, is fed by efficient engineering solutions to real-world problems, under both the constraint and the guidance of bio-inspiration. However, the efficiency and relevance of top-down design for neuromorphic engineering is conditioned by the bio-inspired elements that are considered as essential, with widely different choices reported in Section~\ref{sec_topdown}. This assessment actually bears key importance, yet it is often not sufficiently grounded on theoretical and/or experimental evidence. 

It is worth noting that the bottom-up and top-down approaches discussed in this survey apply only to the \textit{design} of neuromorphic processing systems, and not to how these designs are used in applications. Indeed, bottom-up approaches have some degree of flexibility: they can be used both to understand the computational principles used by the brain and to develop prototypes and testbeds for the deployment of engineering-driven solutions. However, this comes at the cost of a degraded power-performance-area tradeoff compared to their top-down design counterparts (e.g.,~see Fig.~\ref{fig_ppa}), which are typically highly optimized for the use cases they were designed for, and thus less flexible. It directly results from the application being the starting point for top-down designs, while it is the end point for bottom-up ones (Fig.~\ref{fig_strategy}). This highlights the open challenge of achieving application efficiency while maintaining flexibility, which is currently a key driver toward blurring the frontier between purely bottom-up and top-down design approaches. This survey comes at a timely moment to highlight this early convergence, as the first designs merging both standpoints start appearing (Section~\ref{ssec_siac}).

Indeed, both approaches can act as a guide to address the shortcomings of each other (Fig.~\ref{fig_strategy}). On one hand, top-down guidance helps pushing bottom-up neuron and synapse integration beyond the purpose of exploratory neuroscience-oriented experimentation platforms. On the other hand, more bottom-up investigation is needed to identify the computational primitives and mechanisms of the brain that are useful in engineered systems, and to distinguish them from artefacts induced by evolution to compensate for the non-idealities of the biological substrate. The concept of \textit{neuromorphic intelligence} reflects this convergence of natural and artificial intelligence, which requires an integrative view not only of the global approach (i.e.~bottom-up or top-down), but also along the processing chain (i.e.~from sensing to action through computation) and down to the technological design choices outlined in Section~\ref{sec_design}.

\subsection{Open challenges and opportunities}\label{ssec_challenges}

Two key components are still missing to help achieve neuromorphic intelligence and to design neuromorphic systems with a clear competitive advantage against conventional approaches: research and development frameworks, and adequate benchmarks.

\paragraph*{Frameworks}~Unveiling the road to neuromorphic intelligence requires a clearly-articulated framework that should provide three elements. The first element is the definition of appropriate abstraction levels that can be formalized, from the behavior down to the biological primitives. For this, the NEF~\cite{Eliasmith04} and the free energy principle (FEP)~\cite{Friston10} may be good candidates. The former approaches the modeling of complex neural ensembles as dynamical systems of nonlinear differential equations. Support for the NEF is available down to the silicon level with Braindrop~\cite{Neckar19}, which allows mapping dynamical systems onto neuromorphic hardware made of somas and synaptic filters. A large scope of NEF applications has already been studied in the literature (e.g.,~see~\cite{VoelkerThesis19} for a recent review). The latter, the FEP, articulates action, perception and learning into a surprise minimization problem. The FEP has the potential to unify several existing brain theories at different abstraction levels, from the smallest synapse-level scales to network, system, behavioral and evolutionary scales (e.g.,~see~\cite{Kappel21} for a review). The second element required for a framework toward neuromorphic intelligence is a coherent methodology. By reviewing the bottom-up and top-down approaches as well as their strengths, drawbacks, and synergies, this survey provides a first step in this direction. Finally, the framework needs to provide clear metrics and guidelines to measure progress toward neuromorphic intelligence, an aspect that is closely linked to the lack of suitable benchmarks described hereafter. These three framework ingredients bear key importance as recent calls from both industry and academia stress a need for consolidating the field of neuromorphic engineering in a clear direction~\cite{Davies19,Mehonic22}. On top of this three-step framework, a final missing enabler is the support from full software frameworks for streamlining the deployment of neuromorphic applications. One such open-source framework is Lava, which was recently released by Intel together with Loihi 2 and can be extended to support any neuromorphic platform~\cite{Intel21}.

\paragraph*{Benchmarks}~Appropriate benchmarks are missing at two levels. First, \textit{task-level benchmarks} suitable for neuromorphic architectures are required in order to demonstrate an efficiency advantage over conventional approaches. In Section~\ref{sssec_topdown_tradeoff}, while the MNIST dataset was used to highlight that the accuracy/efficiency tradeoff of neuromorphic chips is catching up with state-of-the-art machine-learning accelerators, it was chosen mainly because it is the only dataset currently allowing for such comparisons. Indeed, MNIST does not capture the key dimension inherent to SNNs and neuromorphic computing: time~\cite{Indiveri19}. It is thus unlikely for a neuromorphic efficiency advantage to be demonstrated on MNIST. N-MNIST introduces this time dimension artificially as it is generated with an event-based neuromorphic vision sensor from static images. Moreover, while it is popular for the development of spike-based algorithms and software- or FPGA-based SNNs (e.g.,~see~\cite{Sironi18} for a review), to the best of our knowledge, none of the bottom-up and top-down neuromorphic designs discussed in this survey were benchmarked on N-MNIST, except in~\cite{Frenkel20} for SPOON and in~\cite{Pei19} where Pei~\textit{et~al.}~use this dataset to quantify the efficiency and throughput improvement of Tianjic over GPUs. This further highlights the need for widely-accepted neuromorphic datasets embedding relevant timing information, as recently called for in~\cite{Davies19}. The IBM DVS Gestures dataset~\cite{Amir17} captures 11 classes of hand gestures with an event-based neuromorphic vision sensor and was recently adopted for the benchmarking of large-scale neuromorphic platforms such as TrueNorth and Loihi~\cite{Amir17,Rueckauer22}. However, these platforms relied on deep spiking convolutional networks without exploiting the intrinsic spike timings, as opposed to ReckOn which relied on an RNN topology to solve this task~\cite{Frenkel22}. On the other hand, recent trends in keyword spotting may offer an interesting common task-level benchmark for neuromorphic designs and machine-learning accelerators in the near future. Indeed, the time dimension now becomes an essential component, and spiking auditory sensors can be used on standard datasets such as TIDIGITS or the Google Speech Command Dataset~\cite{Anumula18,Wang21b}, while the recently proposed Heidelberg spiking datasets, which come in \textit{Digits} and \textit{Speech Commands} variants, were generated from a model of the inner ear~\cite{Cramer20}. For the promising use case of biosignal processing (see Section~\ref{ssec_appli}), an electromyography- and vision-based sensor fusion dataset for hand gesture classification was recently proposed in~\cite{CeoliniFrenkelShrestha20}. Data is available in both spiking and non-spiking formats, allowing for fair comparisons between neuromorphic and conventional approaches. Results are already available for an ODIN/MorphIC system, Loihi, and an NVIDIA Jetson Nano portable GPU, showing a favorable accuracy/efficiency tradeoff for the neuromorphic systems. Overall, we would like to emphasize that although demonstrating an advantage for neuromorphic application-specific integrated circuits (ASICs) over general-purpose CPUs and GPUs is a valuable first step, the challenge is now to demonstrate a compelling advantage over conventional machine learning ASICs, such as~\cite{Giraldo20,Shan20} for keyword spotting, and~\cite{Liu21} for biosignal processing tasks.%

Second, \textit{general benchmarks} should also allow for a proper evaluation of neuromorphic intelligence. This assessment cannot be done on specific tasks, as prior task-specific knowledge can be engineered into a system through extended hyperparameter tuning or acquired through massive training data~\cite{Chollet19}. Instead, such benchmarks should measure the end-to-end ability of the system to adapt and generalize, and thus measure its efficiency in acquiring new skills~\cite{Chollet19}. To date, general datasets and task definitions suitable for the assessment of small-scale neuromorphic intelligence are still missing, but an important step toward this goal can be seen in the definition of \textit{closed-loop} benchmarks, where neuromorphic agents have to dynamically sense and act in tight interaction with their environment~\cite{Stewart15,Milde22}.%

Importantly, from task-level to general benchmarks, and from algorithms to systems, the neuromorphic community has recently started driving the \textit{NeuroBench} initiative~\cite{Yik23}, which aims to release a benchmark suite that will allow for fair comparisons across the heterogeneous landscape of neuromorphic approaches and solutions.

\subsection{Neuromorphic applicative landscape: future directions}\label{ssec_appli}
\vspace*{-0.7mm}

The purpose of this section is not to provide an extensive overview of the whole applicative landscape of neuromorphic systems, but rather to outline some of the most promising current and future use cases. These high-potential use cases are mainly at the edge, where low-power resource-constrained devices must process incoming data in an always-on, event-driven fashion. In all of the applications described below, on-chip learning will be a key feature to enable autonomous adaptation to users and environments while ensuring privacy. For neuromorphic applications beyond the scope of adaptive edge computing, we refer the reader to~\cite{Davies21}, which provides a thorough overview based on the Intel Loihi platform.

\paragraph*{Smart sensors}~The use case of smart sensors is currently the dominant one in the literature. As highlighted throughout this survey, it is currently mostly driven by small-scale image recognition. However, as discussed in Section~\ref{ssec_challenges}, keyword spotting embeds biological-time temporal data and may soon be a key driver for neuromorphic smart sensors. Early proof-of-concept works in this direction can be seen in~\cite{Blouw19,Yan21}, though they still rely on keyword spotting datasets that have been pre-processed off-chip to extract the Mel-frequency cepstral coefficient (MFCC) features, which is problematic for two reasons. First, it removes the most computationally-expensive part of the problem (e.g.,~see~\cite{Shan20}). Second, as the MFCC algorithm is anti-causal, it breaks the link between sensory time and processing time, which entails buffering overhead. Therefore, end-to-end time-domain processing of speech data in adaptive neuromorphic smart sensors appears as an exciting direction for future research, with first demonstrations provided with HICANN-X~\cite{Cramer22} (off-chip training) and ReckOn~\cite{Frenkel22} (on-chip training), based on the Heidelberg spiking datasets with raw spike-based data~\cite{Cramer20}.

\paragraph*{Biosignal processing}~Biological signals share with speech two key properties that make them suitable for neuromorphic processing at the edge in wearables and implantables: they involve temporal data and unfold in biological time. Furthermore, biosignals offer the additional advantage of being intrinsically based on a spiking activity, thus allowing for end-to-end spike-based processing. Therefore, there has recently been extensive work on the processing of ExG signals with neuromorphic systems, i.e. electrocardiography (ECG)~\cite{Bauer19,Corradi19}, electroencephalography (EEG)~\cite{Mashford17,SharifshazilehBurelo21}, and electromyography (EMG)~\cite{Donati19,CeoliniFrenkelShrestha20}. Detailed reviews are available in~\cite{Azghadi20,Covi21}. As biosignals are subject to wide variations over time and on a user-to-user basis, on-chip adaptation is also a key requirement~\cite{Covi21}.

\paragraph*{Neuromorphic robots}~The use of neuromorphic processing in robotics is currently actively being investigated~\cite{Sandamirskaya14,Conradt15,Milde17,Kreiser18,Bartolozzi18,Indiveri19,Kreiser20,Zhao20,Yan21}, from closed sensorimotor loops to simultaneous localization and mapping (SLAM), path planning and control. However, importantly, the design of autonomous robotic agents is not only a suitable use case for neuromorphic systems \textit{per se}, but may also be an essential step for bottom-up analysis by synthesis. Indeed, achieving cognition and neuromorphic intelligence \textit{in silico} may not be possible without a body that interacts and adapts continuously with the environment~\cite{Man19,Bartolozzi22}, as it is one of the very purposes biological brains evolved for~\cite{Wolpert11,Anderson16}.%

\section*{Acknowledgment}

The authors would like to thank Jo\~ao Sacramento, Martin Lefebvre, Jean-Didier Legat, Melika Payvand, Yi\u{g}it Demira\u{g}, Elisa Donati, Douwe den Blanken, Lyana Usa, their colleagues at the Institute of Neuroinformatics for fruitful discussions, and the neuromorphic community for e-mail feedback on the first preprint of this work with helpful suggestions.

This work was supported in part by the CHIST-ERA grant CHIST-ERA-18-ACAI-004 (SNSF 20CH21186999 / 1), the European Research Council (ERC) under the European Union's Horizon 2020 research and innovation program grant agreement No 724295, the fonds europ\'een de d\'eveloppement r\'egional FEDER, the Wallonia within the ``Wallonie-2020.EU'' program, the Plan Marshall, and the National Foundation for Scientific Research (F.R.S.-FNRS) of Belgium.

C. Frenkel would particularly like to acknowledge the freedom of research granted by the FNRS during her Ph.D. at UCLouvain as an FNRS Research Fellow.

\begin{IEEEbiography}
	[{\includegraphics[width=1in,height=1.25in,clip,keepaspectratio]{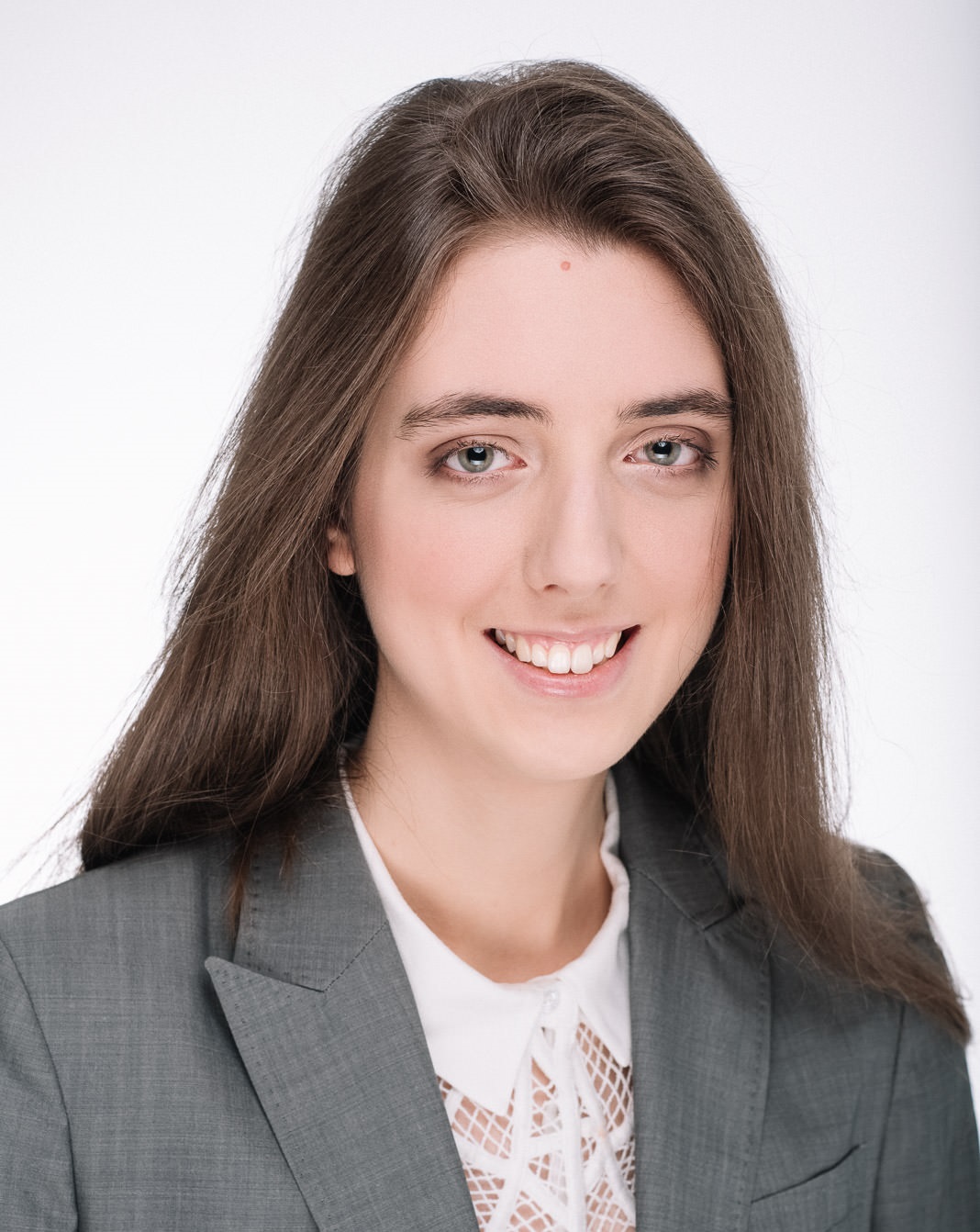}}]{Charlotte Frenkel}
received the M.Sc. degree (\textit{summa cum laude}) in Electromechanical Engineering and the Ph.D. degree in Engineering Science from Universit\'e catholique de Louvain (UCLouvain), Louvain-la-Neuve, Belgium in 2015 and 2020, respectively. In February 2020, she joined the Institute of Neuroinformatics, UZH and ETH Z\"urich, Switzerland, as a postdoctoral researcher. Since July 2022, she is an Assistant Professor at Delft University of Technology, The Netherlands.

Her current research aims at bridging the bottom-up and top-down design approaches toward neuromorphic edge intelligence, with a focus on spiking neural network processor design, embedded machine learning, and on-chip training algorithms.

Ms. Frenkel received a best paper award at the IEEE Intl. Symp. on Circuits and Systems (ISCAS) 2020 and, for her Ph.D. thesis, the FNRS Nokia Bell Labs Scientific Award 2021, the FNRS IBM Innovation Award 2021, and the UCLouvain/ICTEAM Best Thesis Award 2021. In 2023, she received a prestigious AiNed Fellowship Grant from the Dutch Research Council (NWO). She serves as a TPC member for the tinyML Research Symposium, the IEEE European Solid-State Circuits Conference (ESSCIRC), the IEEE Intl. Symp. on Low-Power Electronics and Design (ISLPED), and the IEEE Design, Automation and Test in Europe (DATE) conference since 2022, as a member of the neuromorphic systems and architecture technical committee of the IEEE CAS society since 2021, as an associate editor for the IEEE Trans. on Biomedical Circuits and Systems, and as a reviewer for various conferences and journals, including the IEEE J.~of Solid-State Circuits, IEEE Trans. on Neural Networks and Learning Syst., IEEE Trans. on Circuits and Syst. I/II, Nature, Nature Electronics, and Nature Machine Intelligence. She presented several invited talks, including two keynotes at the tinyML EMEA technical forum 2021 and at the Neuro-Inspired Computational Elements (NICE) neuromorphic conference 2021. She is the chair of the tinyML initiative on neuromorphic engineering and a program co-chair of the NICE conference 2023.
\end{IEEEbiography}

\begin{IEEEbiography}
	[{\includegraphics[width=1in,height=1.25in,clip,keepaspectratio]{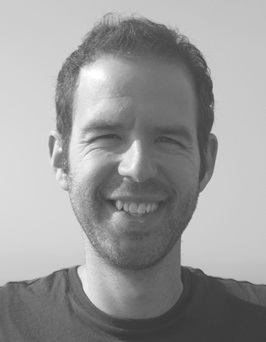}}]{David Bol}
received the Ph.D degree in Engineering Science from Universit\'e catholique de Louvain (UCLouvain), Louvain-la-Neuve, Belgium in 2004 and 2008, respectively. In 2005, he was a visiting Ph.D student at the CNM National Centre for Microelectronics, Sevilla, Spain, in advanced logic design. In 2009, he was a postdoctoral researcher at intoPIX, Louvain-la-Neuve, Belgium, in low-power design for JPEG2000 image processing. In 2010, he was a visiting postdoctoral researcher at the UC Berkeley Laboratory for Manufacturing and Sustainability, Berkeley, CA, in life-cycle assessment of the semiconductor environmental impact. He is now an Associate Professor at UCLouvain. In 2015, he participated to the creation of e-peas semiconductors, Louvain-la-Neuve, Belgium.

Prof. Bol leads the Electronic Circuits and Systems (ECS) group focused on ultra-low-power design of integrated circuits for environmental and biomedical IoT applications including computing, power management, sensing and wireless communications with a holistic focus on environmental sustainability. He is actively engaged in a social-ecological transition in the field of ICT research with a post-growth approach.

Prof. Bol has authored or co-authored more than 150 technical papers and conference contributions and holds three delivered patents. He (co-)received four Best Paper/Poster/Design Awards in IEEE conferences (ICCD 2008, SOI Conf. 2008, FTFC 2014, ISCAS 2020). He served as an editor for MDPI J. Low-Power Electronics and Applications, as a TPC member of IEEE SubVt/S3S conferences and currently serves as a reviewer for various journals and conferences such as IEEE J. of Solid-State Circuits, IEEE Trans. on VLSI Syst., IEEE Trans. on Circuits and Syst. I/II. Since 2008, he regularly presents invited papers and keynote tutorials in international conferences including invited~\mbox{forum presentations at IEEE ISSCC 2018 and IEEE Symp. VLSI 2022.}
\hspace*{3mm}On the private side, Prof. Bol pioneered the parental leave for male professors in his faculty, to spend time connecting to nature with his family.
\end{IEEEbiography}

\begin{IEEEbiography}
	[{\includegraphics[width=1in,height=1.25in,clip,keepaspectratio]{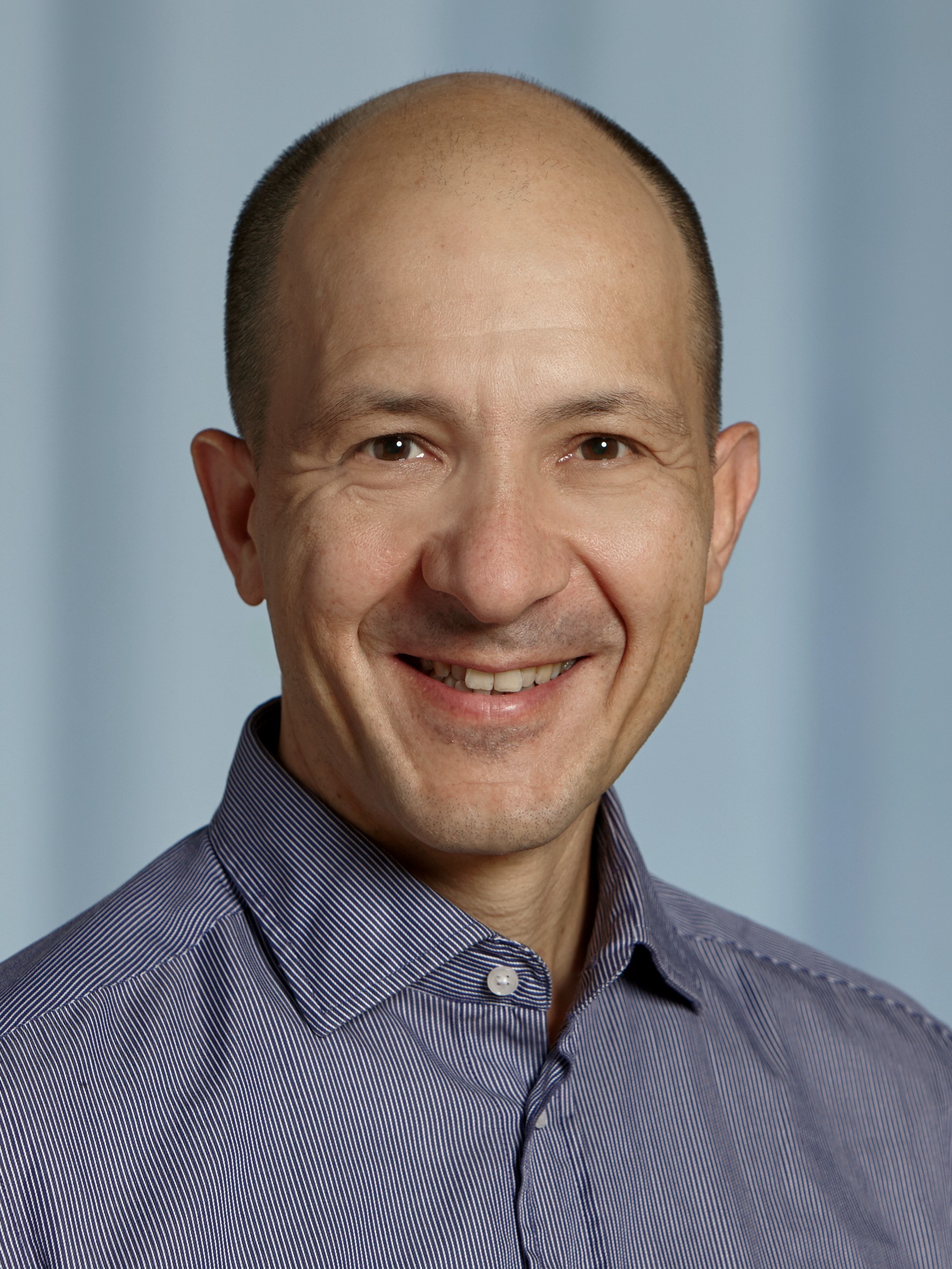}}]{Giacomo Indiveri}
is a dual Professor at the Faculty of Science of the University of Zurich and at Department of Information Technology and Electrical Engineering of ETH Zurich, Switzerland. He is the director of the Institute of Neuroinformatics of the University of Zurich and ETH Zurich. He obtained an M.Sc. degree in electrical engineering in 1992 and a Ph.D. degree in computer science from the University of Genoa, Italy in 2004. 
Engineer by training, Indiveri has also expertise in neuroscience, computer science, and machine learning. He has been combining these disciplines by studying natural and artificial intelligence in neural processing systems and in neuromorphic cognitive agents. His latest research interests lie in the study of spike-based learning mechanisms and recurrent networks of biologically plausible neurons, and in their integration in real-time closed-loop sensory-motor systems designed using analog/digital circuits and emerging memory technologies. His group uses these neuromorphic circuits to validate brain inspired computational paradigms in real-world scenarios, and to develop a new generation of fault-tolerant event-based neuromorphic computing technologies. Indiveri is senior member of the IEEE society, and a recipient of the 2021 IEEE Biomedical Circuits and Systems Best Paper Award. He is also an ERC fellow, recipient of three European Research Council grants.
\end{IEEEbiography}

\end{document}